\documentclass[10pt,twocolumn,letterpaper]{article}

\usepackage[pagenumbers]{cvpr} 

\usepackage{graphicx}
\usepackage{amsmath}
\usepackage{amssymb}
\usepackage{booktabs}
\DeclareMathOperator*{\argmax}{\arg\!\max}
\usepackage[table]{xcolor}
\definecolor{bestcolor}{rgb}{1.0,1.0,1.0} 
\definecolor{secondcolor}{rgb}{1.0,1.0,1.0} 
\definecolor{thirdcolor}{rgb}{1.0,1.0,1.0} 
\newcommand{\bone}{\cellcolor{bestcolor}}
\newcommand{\btwo}{\cellcolor{secondcolor}}
\newcommand{\bthird}{\cellcolor{thirdcolor}}
\definecolor{initcolor}{rgb}{.830,.550,.400} 
\definecolor{optcolor}{rgb}{.400,0.490,.820} 

\usepackage[pagebackref,breaklinks,colorlinks]{hyperref}

\usepackage{url}            
\usepackage{booktabs}       
\usepackage{amsfonts}       
\usepackage{times}
\usepackage{epsfig}
\usepackage{amsmath}
\usepackage{amssymb}
\usepackage{multirow}
\usepackage{subcaption}
\usepackage{cuted}
\usepackage{changepage}
\usepackage{amsthm}
\usepackage{xcolor}
\usepackage{enumitem}

\usepackage[capitalize]{cleveref}
\crefname{section}{Sec.}{Secs.}
\Crefname{section}{Section}{Sections}
\Crefname{table}{Table}{Tables}
\crefname{table}{Tab.}{Tabs.}

\def\cvprPaperID{3358} 
\def\confName{CVPR}
\def\confYear{2022}
\usepackage{appendix}

\newcommand{\pnerf}{Point-NeRF}
\newcommand{\boldstart}[1]{\noindent\textbf{#1}}
\newcommand{\boldstartspace}[1]{\vspace{0.05in}\noindent\textbf{#1}}

\begin{document}

\title{Point-NeRF: Point-based Neural Radiance Fields}

\author{Qiangeng Xu$^{1}$ $^\dagger$ \qquad Zexiang Xu$^{2}$ \qquad Julien Philip 
$^{2}$ \qquad Sai Bi$^{2}$  \qquad Zhixin Shu$^{2}$ 
\\  Kalyan Sunkavalli$^{2}$ \qquad \qquad Ulrich Neumann$^{1}$ \\
    \hspace{-15mm}$^1$University of Southern California \hspace{30mm} $^2$Adobe Research\\
    {\tt\small \hspace{0mm}\{qiangenx,uneumann\}@usc.edu}\hspace{5mm}{\tt\small \{zexu,juphilip,sbi,zshu,sunkaval\}@adobe.com}\qquad
}

\makeatletter
\let\@oldmaketitle\@maketitle
\renewcommand{\@maketitle}{\@oldmaketitle

\begin{center}
    \begin{adjustwidth}{0pt}{0pt}
        \centering
        \includegraphics[width=0.99\textwidth]{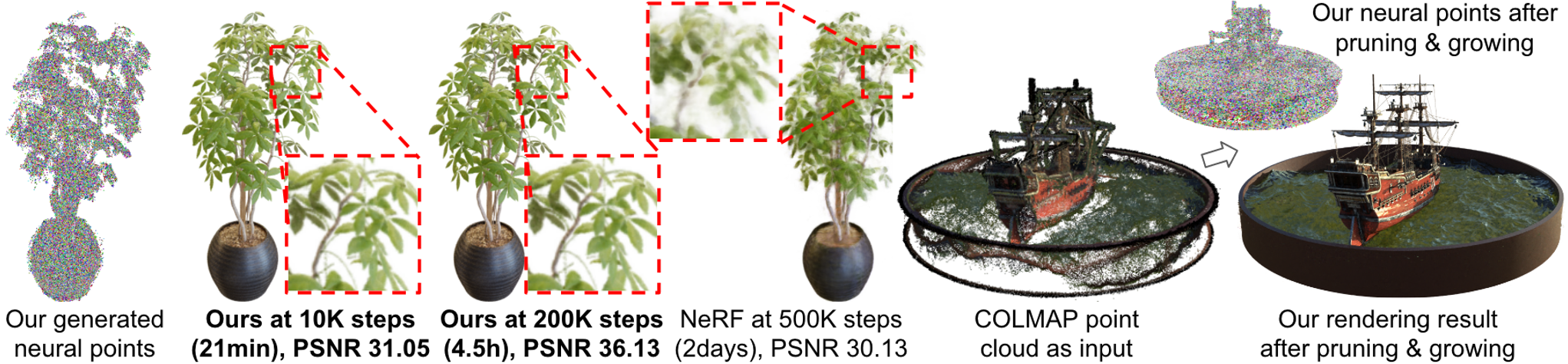}
        \captionof{figure}{Point-NeRF uses neural 3D points to efficiently represent and render a continuous radiance volume. The point-based radiance field can be predicted via network forward inference from multi-view images. It can then be optimized per scene to achieve reconstruction quality that surpasses NeRF \cite{mildenhall2020nerf} in tens of minutes. Point-NeRF can also leverage off-the-shelf reconstruction methods like COLMAP \cite{schoenberger2016mvs} and is able to perform point pruning and growing that automatically fix the holes and outliers that are common in these approaches.}
        \label{fig:teaser}
    \end{adjustwidth}
\end{center}
 }
\makeatother

\maketitle

\let\thefootnote\relax\footnotetext{\leftline{$^\dagger$This work is partially done during the internship at Adobe Research.}}
\begin{abstract}
Volumetric neural rendering methods like NeRF \cite{mildenhall2020nerf} generate high-quality view synthesis results but are optimized per-scene leading to prohibitive reconstruction time. On the other hand, deep multi-view stereo methods can quickly reconstruct scene geometry via direct network inference. \pnerf{} combines the advantages of these two approaches by using neural 3D point clouds, with associated neural features, to model a radiance field. \pnerf{} can be rendered efficiently by aggregating neural point features near scene surfaces, in a ray marching-based rendering pipeline. Moreover, \pnerf{} can be initialized via direct inference of a pre-trained deep network to produce a neural point cloud; this point cloud can be finetuned to surpass the visual quality of NeRF with $30\times$ faster training time. \pnerf{} can be combined with other 3D reconstruction methods and handles the errors and outliers in such methods via a novel pruning and growing mechanism. The experiments on the DTU \cite{dtu}, the NeRF Synthetics \cite{mildenhall2020nerf}, the ScanNet \cite{dai2017scannet} and the Tanks and Temples \cite{Knapitsch2017} datasets demonstrate \pnerf{} can surpass the existing methods and achieve the state-of-the-art results. Please visit our website \url{https://xharlie.github.io/projects/project_sites/pointnerf} for code and more results.
\end{abstract}
\section{Introduction}
    
    
 
Modeling real scenes from image data and rendering photo-realistic novel views is a central problem in computer vision and graphics.
NeRF \cite{mildenhall2020nerf} and its extensions \cite{liu2020neural,martin2021nerf,zhang2020nerf++} have shown great success on this by modeling neural radiance fields.
These methods \cite{mildenhall2020nerf,zhang2020nerf++,park2021nerfies} often reconstruct radiance fields using global MLPs for the entire space through ray marching.
This leads to long reconstruction times due to the slow per-scene network fitting and the unnecessary sampling of vast empty space.

We address this issue using Point-NeRF, a novel point-based radiance field representation that uses 3D neural points to model a continuous volumetric radiance field. Unlike NeRF that purely depends on per-scene fitting, Point-NeRF can be effectively initialized via a feed-forward deep neural network, pre-trained across scenes. Moreover, Point-NeRF avoids ray sampling in the empty scene space by leveraging classical point clouds that approximate the actual scene geometry. This advantage of Point-NeRF leads to more efficient reconstruction and more accurate rendering than other neural radiance field models~\cite{mildenhall2020nerf,chen2021mvsnerf,ibrnet,yu2020pixelnerf}.



Our Point-NeRF representation consists of a point cloud with per-point neural features: each neural point encodes the local 3D scene geometry and appearance around it. Prior point-based rendering techniques \cite{aliev2020neural} use similar neural point clouds but perform rendering with rasterization and 2D CNNs operating in image space. We instead treat these neural points as local neural basis functions in 3D to model a continuous volumetric radiance field which enables high-quality rendering using differentiable ray marching. In particular, for any 3D location, we propose to use an MLP network to aggregate the neural points in its neighborhood to regress the volume density and view-dependent radiance at that location. This expresses a continuous radiance field.

We present a learning-based framework to efficiently initialize and optimize the point-based radiance fields. To generate a initial field, we leverage deep multi-view stereo (MVS) techniques \cite{yao2018mvsnet}, i.e., applying a cost-volume-based network to predict depth which is then unprojected to 3D space. In addition, a deep CNN is trained to extract 2D feature maps from input images, naturally providing the per-point features. These neural points from multiple views are combined as a neural point cloud, which forms a point-based radiance field of the scene. We train this point generation module with the point-based volume rendering networks from end to end, to render novel view images and supervise them with the ground truth. This leads to a \emph{generalizable} model that can directly predict a point-based radiance field at inference time. Once predicted, the initial point-based field is further optimized per scene in a short period to achieve photo-realistic rendering. As shown in Fig.~\ref{fig:teaser} (left), 21 minutes of optimization with Point-NeRF outperforms a NeRF model trained for days. 

Besides using the in-built point cloud reconstruction, our approach is \emph{generic} and can also generate a radiance field based on a point cloud of other reconstruction techniques. However, the reconstructed point cloud produced by techniques like COLMAP \cite{schoenberger2016mvs}, in practice, contain holes and outliers that adversely affect the final rendering. To address this issue, we introduce \emph{point growing} and \emph{pruning} as part of our optimization process. We leverage the geometric reasoning during volume rendering \cite{drebin1988volume} and grow points near the point cloud boundary in high volume density regions and prune points in low-density regions. The mechanism effectively improves our final reconstruction and rendering quality. We show an example in Fig.~\ref{fig:teaser} (right) where we convert COLMAP points to a radiance field and successfully fill large holes and produce photo-realistic renderings.


We train our model on the DTU dataset \cite{dtu} and evaluate on DTU testing scenes, NeRF synthetic, Tanks \& Temples~\cite{Knapitsch2017}, and ScanNet~\cite{dai2017scannet} scenes. The results demonstrate that our approach can achieve state-of-the-art novel view synthesis, outperforming many prior arts including point-based methods \cite{aliev2020neural}, NeRF, NSVF \cite{liu2020neural}, and many other generalizable neural methods \cite{yu2020pixelnerf,ibrnet,chen2021mvsnerf} (see (Tab.~\ref{tb:dtu} and \ref{tb:nerfsynth})). 


\section{Related Work}

\boldstart{Scene representations.}
Traditional and neural methods have studied many 3D scene representations, including volumes \cite{seitz1999photorealistic,kutulakos2000theory,ji2017surfacenet,wu20153d,qi2016volumetric}, point clouds \cite{qi2017pointnet,achlioptas2018learning,wang2018mvpnet}, meshes \cite{kanazawa2018learning,wang2018pixel2mesh}, depth maps \cite{liu2015learning,huang2018deepmvs}, and implicit functions \cite{chen2018learning,mescheder2018occupancy,niemeyer2020differentiable,yariv2020multiview}, in diverse vision and graphics applications.
Recently, various neural scene representations have been presented \cite{zhou2018stereo,sitzmann2019deepvoxels,lombardi2019neural,bi2020deep}, advancing the state of the art in novel view synthesis and realistic rendering, with volumetric neural radiance fields (NeRFs) \cite{mildenhall2020nerf} producing high fidelity results. 
NeRFs are often reconstructed as global MLPs \cite{mildenhall2020nerf,zhang2020nerf++,park2021nerfies} that encode the entire scene space; this can be inefficient and expensive when reconstructing complex and large-scale scenes.
Instead, Point-NeRF is a localized neural representation, combining volumetric radiance fields with point clouds that are classically used to approximate scene geometry.
We distribute fine-grained neural points to model complex local scene geometry and appearance, leading to better rendering quality than NeRF 
(see Fig.~\ref{fig:dtu},~\ref{fig:nerfsynth}).

Voxel grids with per-voxel neural features \cite{liu2020neural,chen2021mvsnerf,hedman2021baking} are also a local neural radiance representation. However, our point-based representation adapts better to actual surfaces, leading to better quality. Also, we directly predict good initial neural point features, bypassing the per-scene optimization that is required by most voxel-based methods \cite{liu2020neural,hedman2021baking}.

\boldstartspace{Multi-view reconstruction and rendering.}
Multi-view 3D reconstruction has been extensively studied and addressed with a number of structure-from-motion \cite{schoenberger2016sfm,vijayanarasimhan2017sfm,tang2018ba} and multi-view stereo techniques \cite{furukawa2009accurate,kutulakos2000theory,schoenberger2016mvs,yao2018mvsnet,cheng2020deep}.
Point clouds are often the direct output from MVS or depth sensor, though they are usually converted to meshes \cite{lorensen1987marching,kazhdan2006poisson} for rendering and visualization.
Meshing can introduce errors and may require image-based rendering \cite{debevec1998efficient,buehler2001unstructured,zhou2014color} for high-quality rendering.
We instead directly use point clouds from deep MVS to achieve realistic rendering.

Point clouds have been widely used in rendering, often via rasterization-based point splatting, and even differentiable rasterization modules \cite{wiles2020synsin,lassner2021pulsar}.
However, reconstructed point clouds often have holes and outliers that lead to artifacts in rendering. Point-based neural rendering methods address this by splatting neural features and using 2D CNNs to render them \cite{aliev2020neural,kopanas2021point,meshry2019neural}.
In contrast, our point-based approach utilizes 3D volume rendering,  
leading to significantly better results than previous point-based methods. 

\boldstartspace{Neural radiance fields.} NeRFs \cite{mildenhall2020nerf} have demonstrated remarkably high-quality results for novel view synthesis. 
They have been extended to achieve dynamic scene capture \cite{li2021neural,park2021hypernerf}, relighting \cite{bi2020neural,boss2021nerd}, appearance editing \cite{xiang2021neutex}, fast rendering \cite{hedman2021baking,yu2021plenoctrees}, and generative models \cite{chan2021pi,schwarz2020graf,niemeyer2021giraffe}.  
However, most methods \cite{li2021neural,park2021hypernerf,xiang2021neutex,bi2020neural} still follow the original NeRF framework and train per-scene MLPs to represent radiance fields.
We make use of neural points with spatially varying neural features in a scene to encode its radiance field.
This localized representation can model more complex scene content than pure MLPs that have limited network capacity. More importantly, we show that our point-based neural field can be efficiently initialized via a pre-trained deep neural network that generalizes across scenes and leads to highly efficient radiance field reconstruction.

Prior works also present generalizable radiance field-based methods.
PixelNeRF \cite{yu2020pixelnerf} and IBRNet \cite{ibrnet} aggregate multi-view 2D image features at every sampled ray point to regress volume rendering properties for radiance field rendering.
In contrast, we leverage features in 3D neural points around the scene surface to model radiance fields. This avoids sampling points in the vast empty space and leads to higher rendering quality and faster radiance field reconstruction than PixelNeRF and IBRNet.
MVSNeRF \cite{chen2021mvsnerf} can achieve very fast voxel-based radiance field reconstruction. However, its prediction network requires a fixed number of three small-baseline images as input and thus can only efficiently reconstruct local radiance fields.
Our approach can fuse neural points from an arbitrary number of views and achieve fast reconstruction of complete 360 radiance fields which MVSNeRF cannot support. 


\newcommand{\Img}{I}
\newcommand{\Cam}{\Phi}
\newcommand{\iI}{q}
\newcommand{\ImgNum}{Q}

\newcommand{\Color}{c}
\newcommand{\ShadX}{x}
\newcommand{\iS}{j}
\newcommand{\iSS}{t}
\newcommand{\ShadNum}{M}

\newcommand{\Dir}{d}
\newcommand{\Rad}{r}
\newcommand{\Trans}{\tau}
\newcommand{\Dens}{\sigma}
\newcommand{\Step}{\Delta}
\newcommand{\Contr}{\alpha}
\newcommand{\Dist}{\epsilon} 

\newcommand{\PointNum}{N}
\newcommand{\PointX}{p}
\newcommand{\PointCloud}{P}

\newcommand{\iP}{i}
\newcommand{\PointF}{f}
\newcommand{\PointConf}{\gamma}
\newcommand{\PointWeight}{w}

\newcommand{\PointRadius}{R}

\newcommand{\NetPoint}{F}
\newcommand{\NetDense}{T}
\newcommand{\NetRad}{R}
\newcommand{\NetPointPC}{G_{p,\gamma}}
\newcommand{\NetPointF}{G_{f}}

\newcommand{\ThresContr}{T_{\text{opacity}}}
\newcommand{\ThresDist}{T_{\text{dist}}}

\begin{figure*}[!hbt]
        \vspace{-15pt}
        \centering
        \setlength{\abovedisplayskip}{0pt}%
        \setlength{\abovedisplayshortskip}{\abovedisplayskip}%
        \setlength{\belowdisplayskip}{0pt}%
            \begin{subfigure}{0.40\linewidth}
                \centering
                \includegraphics[width=0.95\linewidth]{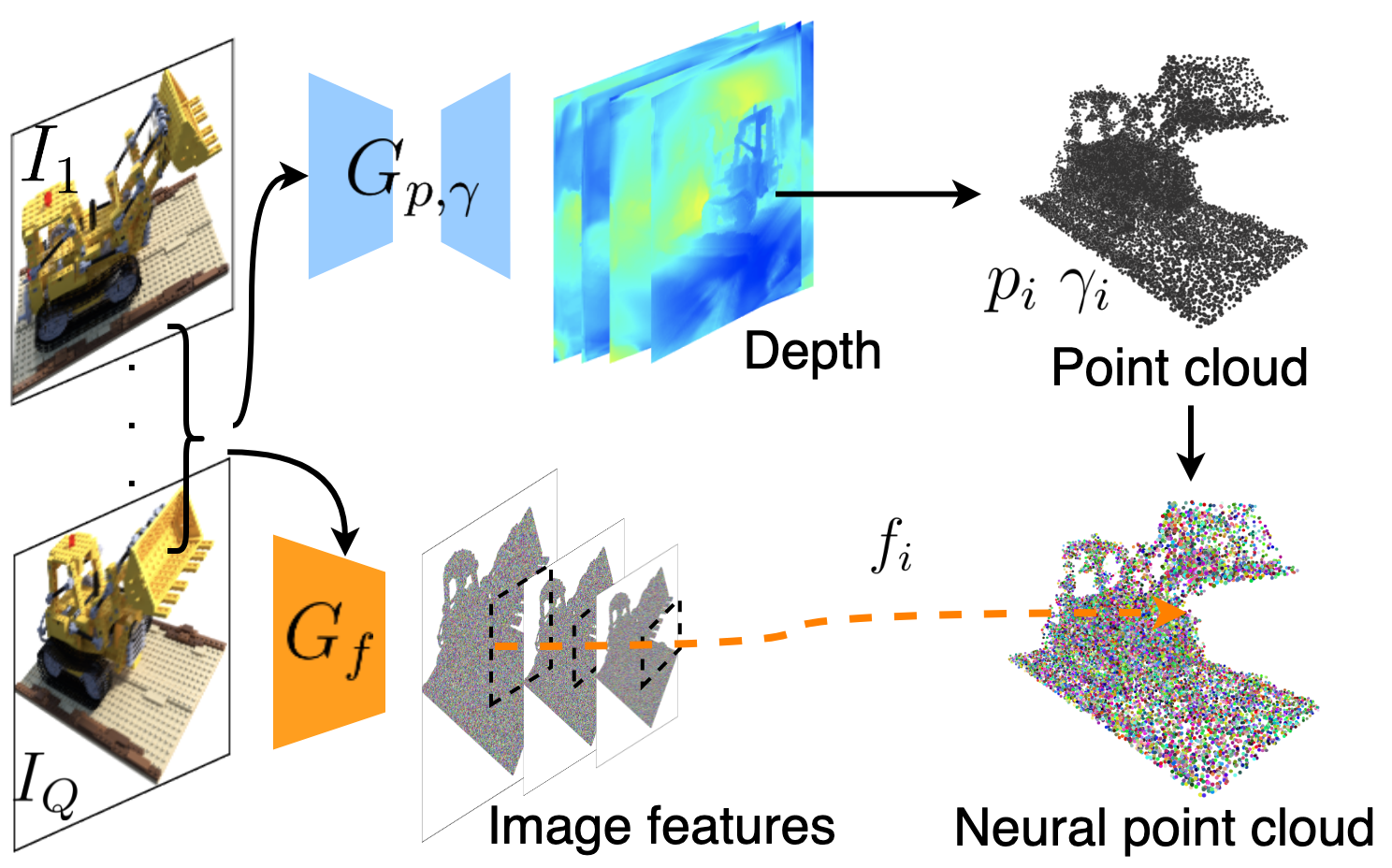}
                \caption{Neural Point Generation.}
            \end{subfigure} \hfill
            \begin{subfigure}{0.55\linewidth}
                \centering
                \includegraphics[width=0.95\linewidth]{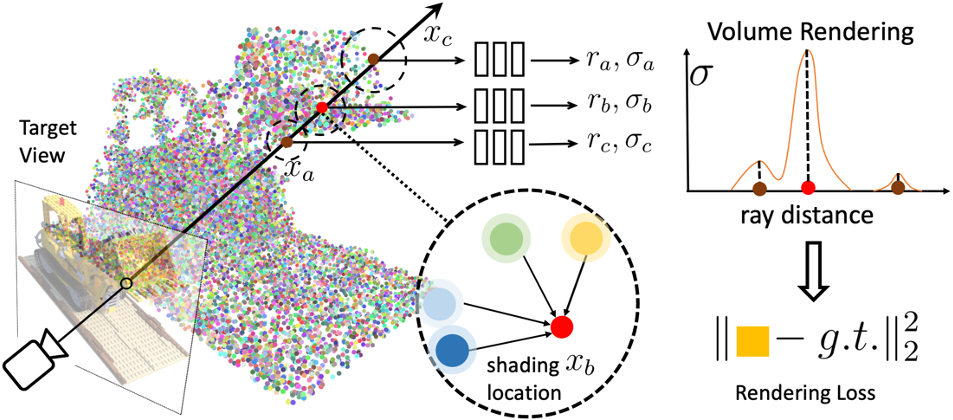} 
                \caption{Point-NeRF Representation with Volume Rendering.}
            \end{subfigure}
        \captionsetup{aboveskip=3pt}
        \captionsetup{belowskip=-10pt}
        \caption{Overview of \pnerf{}. (a) From multi-view images, our model generates depth for each view by using a cost volume-based 3D CNNs $G_{p,\gamma}$ and extract 2D features from the input images by a 2D CNN $G_f$. After aggregating the depth map, we obtain a point-based radiance field in which each point has a spatial location $p_i$, a confidence $\gamma_i$ and the unprojected image features $f_i$. (b) To synthesize a novel view, we conduct differentiable ray marching and compute shading only nearby the neural point cloud (e.g., $x_a, x_b, x_c$). At each shading location, \pnerf{} aggregates features from its K neural point neighbors and compute radiance $r$ and volume density $\sigma$ then accumulate $r$ using $\sigma$. The entire process is end-to-end trainable and the point-based radiance field can be optimized with the rendering loss.}
        \label{fig:overview}
    \end{figure*}
    
    
\section{Point-NeRF Representation}
\label{sec:pcl_render}
We present our novel point-based radiance field representation, designed for efficient reconstruction and  rendering (see Fig. \ref{fig:overview} (b)). We start with some preliminaries.

\boldstartspace{Volume rendering and radiance fields.}
Physically-based volume rendering can be numerically evaluated via differentiable ray marching. Specifically, a pixel's radiance can be computed by marching a ray through the pixel, sampling $\ShadNum$ shading points at $\{\ShadX_\iS\;|\;\iS=1,...,\ShadNum\}$ along the ray, and accumulating radiance using volume density, as:
\begin{adjustwidth}{0pt}{0pt}
    \setlength{\abovedisplayskip}{-5pt}%
    \setlength{\abovedisplayshortskip}{\abovedisplayskip}%
    \setlength{\belowdisplayshortskip}{-5pt}%
    \begin{equation}
    \begin{aligned}
        \Color &=  \sum_\ShadNum \Trans_\iS (1-\exp (-\Dens_\iS \Step_\iS)) \Rad_\iS, \\
        \Trans_\iS &= \exp (-\sum_{\iSS=1}^{\iS-1} \Dens_\iSS \Step_\iSS ). 
    \label{eq:raymarching}
    \end{aligned}
    \end{equation}
\end{adjustwidth}
Here, $\Trans$ represents volume transmittance; $\Dens_\iS$ and $\Rad_\iS$ are the volume density and radiance for each shading point $\iS$ at $\ShadX_\iS$, $\Step_\iSS$ is the distance between adjacent shading samples.

A radiance field represents the volume density $\Dens$ and view-dependent radiance $\Rad$ at any 3D location.
NeRF \cite{mildenhall2020nerf} proposes to use a multi-layer perceptron (MLP) to regress such radiance fields.
We propose Point-NeRF that instead utilizes a neural point cloud to compute the volume properties, allowing for faster and higher-quality rendering.

\boldstartspace{Point-based radiance field.}
We denote a neural point cloud by $\PointCloud = \{(\PointX_\iP, \PointF_\iP, \PointConf_\iP)|\iP=1,...,\PointNum\}$, where each point $\iP$ is located at $\PointX_\iP$ and associated with a neural feature vector $\PointF_\iP$ that encodes the local scene content. We also assign each point a scale confidence value $\PointConf_\iP \in [0,1]$ that represents how likely that point is being located near an actual scene surface. We regress the radiance field from this point cloud.


Given any 3D location $\ShadX$, we query $K$ neighboring neural points around it within a certain radius $\PointRadius$. Our point-based radiance field can be abstracted as a neural module that regresses volume density $\Dens$ and view-dependent radiance $\Rad$ (along any viewing direction $\Dir$) at any shading location $\ShadX$ from its neighboring neural points as:
\begin{adjustwidth}{-5pt}{0pt}
    \setlength{\abovedisplayskip}{-5pt}%
    \setlength{\belowdisplayshortskip}{-5pt}%
    \begin{equation}
    (\Dens, \Rad) = \text{Point-NeRF}(\ShadX, \Dir, \PointX_1,\PointF_1, \PointConf_1,..., \PointX_K,\PointF_K, \PointConf_K).
    \end{equation}
\end{adjustwidth}
We use a PointNet-like \cite{qi2017pointnet} neural network, with multiple sub-MLPs, to do this regression. Overall, we first conduct neural processing for each neural point and then aggregate the multi-point information to obtain the final estimates.

\boldstartspace{Per-point processing.} We use an MLP $\NetPoint$ to process each neighboring neural point to predict a new feature vector for the shading location $\ShadX$ by:
\begin{adjustwidth}{0pt}{0pt}
    \setlength{\abovedisplayskip}{-5pt}%
    \setlength{\abovedisplayshortskip}{\abovedisplayskip}%
    \begin{equation}
    \PointF_{\iP,\ShadX} = \NetPoint(\PointF_\iP, \ShadX-\PointX_\iP). 
    \end{equation}
\end{adjustwidth}
Essentially, the original feature $\PointF_\iP$ encodes the local 3D scene content around $\PointX_\iP$. 
This MLP network expresses a local 3D function that outputs the specific neural scene description $\PointF_{\iP,\ShadX}$ at $\ShadX$, modeled by the neural point in its local frame. 
The usage of relative position $\ShadX-\PointX$ makes the network invariant to point translation for better generalization. 

\boldstartspace{View-dependent radiance regression.}
We use standard inverse distance weighting to aggregate the neural features $\PointF_{\iP,\ShadX}$ regressed from these K neighboring points to obtain a single feature $\PointF_{\ShadX}$ that describes scene appearance at $\ShadX$:
\begin{adjustwidth}{0pt}{0pt}
    \setlength{\abovedisplayskip}{-5pt}%
    \setlength{\abovedisplayshortskip}{\abovedisplayskip}%
    \begin{equation}
    \PointF_{\ShadX} = \sum_{\iP}  \PointConf_\iP \frac{\PointWeight_\iP}{\sum \PointWeight_\iP} \PointF_{\iP,\ShadX},   \text{where } \PointWeight_\iP=\frac{1}{\|\PointX_\iP-\ShadX\|}.
    \end{equation}
\end{adjustwidth}
Then an MLP, $\NetRad$, regress the view-dependent radiance from this feature given a viewing direction, $\Dir$: 
\begin{adjustwidth}{0pt}{0pt}
    \setlength{\abovedisplayskip}{-5pt}%
    \setlength{\abovedisplayshortskip}{\abovedisplayskip}%
    \begin{equation}
    \Rad = \NetRad(\PointF_\ShadX, \Dir).
    \end{equation}
\end{adjustwidth}
The inverse-distance weight $\PointWeight_\iP$ is widely used in scattered data interpolation; we leverage it to aggregate neural features, making closer neural points contribute more to the shading computation.
In addition, we use the per-point confidence $\PointConf$ in this process; this is optimized in the final reconstruction with a sparsity loss, giving the network the flexibility of rejecting unnecessary points.

\boldstartspace{Density regression.}
To compute volume density $\Dens$ at $\ShadX$, we follow a similar multi-point aggregation.
However, we first regress a density $\Dens_\iP$ per point using an MLP $\NetDense$ and then do inverse distance-based weighting, given by:
\begin{adjustwidth}{0pt}{0pt}
    \setlength{\abovedisplayskip}{-5pt}%
    \setlength{\abovedisplayshortskip}{\abovedisplayskip}%
    \begin{align}
    \Dens_{\iP} &= \NetDense(\PointF_{\iP,\ShadX})\\
    \Dens &= \sum_{\iP} \Dens_\iP \PointConf_\iP \frac{\PointWeight_\iP}{\sum \PointWeight_\iP}, \PointWeight_\iP=\frac{1}{\|\PointX_\iP-\ShadX\|}. \label{eqn:weighted_density}
    \end{align}
\end{adjustwidth}
\vspace{-5pt}
Thus, each neural point directly contributes to the volume density, and point confidence $\PointConf_\iP$ is explicitly associated with this contribution. We leverage this in our point removal process (see Sec.~\ref{sec:prune_grow}).

\boldstartspace{Discussion.} 
Unlike previous neural point-based methods \cite{aliev2020neural,meshry2019neural} that rasterize point features and then render them with 2D CNNs, our representation and rendering are entirely in 3D. 
By using a point cloud that approximates the scene geometry, our representation naturally and efficiently adapts to scene surfaces and avoids sampling shading locations in empty scene space. For shading points along each ray, we implement an efficient algorithm to query neighboring neural points; details are in the supplemental material.

\section{Point-NeRF Reconstruction}
    
\begin{figure}[!hbt]
        \begin{adjustwidth}{0pt}{0pt}
            \setlength{\abovedisplayskip}{0pt}%
            \setlength{\abovedisplayshortskip}{\abovedisplayskip}%
            \setlength{\belowdisplayskip}{0pt}%
            \begin{center}
                \includegraphics[width=1\linewidth]{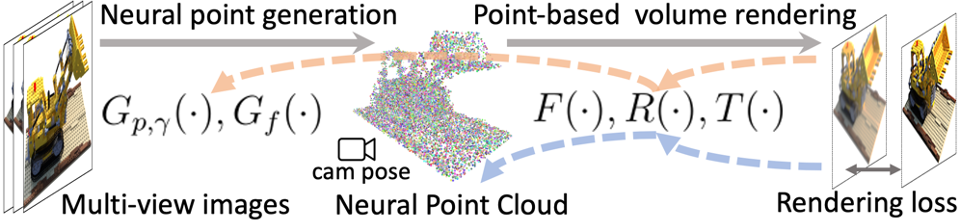}
            \end{center}
        \end{adjustwidth}
        \captionsetup{aboveskip = 2pt}
        \captionsetup{belowskip = -15pt}
        \caption{The dash lines indicate gradient updates for \textcolor{initcolor}{\textbf{radiance field initialization}} and \textcolor{optcolor}{\textbf{per-scene optimization}}.}
        \label{fig:pipeline}
    \end{figure}

We now introduce our pipeline for efficiently reconstructing point-based radiance fields.
We first leverage a deep neural network, trained across scenes, to generate an initial point-based field via direct network inference (Sec.~\ref{sec:pcl_gen}).
This initial field is further optimized per scene with our point growing and pruning techniques, leading to our final high-quality radiance field reconstruction (Sec.~\ref{sec:prune_grow}). Figure.~\ref{fig:pipeline} shows this workflow with the corresponding gradient updates for the initial prediction and per-scene optimization.
    
\subsection{Generating initial point-based radiance fields}
    \label{sec:pcl_gen}
Given a set of known images $\Img_1$,...,$\Img_\ImgNum$, and a point cloud, our Point-NeRF representation can be reconstructed by optimizing the randomly initialized per-point neural features and the MLPs with a rendering loss (similar to NeRF).
However, this pure per-scene optimization depends on an exisiting point cloud, and can be prohibitively slow.
Therefore, we propose a neural generation module to predict all neural point properties, including point locations $\PointX_\iP$, neural features $\PointF_\iP$ and point confidence $\PointConf_\iP$, via a feed-forward neural network for efficient reconstruction. The direct inference of the network outputs a good initial point-based radiance field. 
The initial fields can then be fine-tuned to achieve high-quality rendering. In a very short period, the rendering quality is better or on par with NeRF which takes substantially longer time to optimize (see Tab.~\ref{tb:dtu} and ~\ref{tb:nerfsynth}).

\boldstartspace{Point location and confidence.} We leverage deep MVS methods to generate 3D point locations using cost volume-based 3D CNNs \cite{yao2018mvsnet,cheng2020deep}. Such networks produce high-quality dense geometry and generalize well across domains. For each input image $\Img_\iI$ with camera parameters $\Cam_\iI$ at viewpoint $\iI$, we follow MVSNet \cite{huang2018deepmvs} to first build a plane-swept cost volume by warping 2D image features from neighboring viewpoints and then regress depth probability volume using deep 3D CNNs. A depth map is computed by linearly combining per-plane depth values weighted by the probabilities. We unprojected the depth map to 3D space to get a point cloud $\{\PointX_{1},..., \PointX_{\PointNum_\iI}\}$ per view $\iI$.

Since the depth probabilities describe the likelihood of the point being on the surface, we tri-linearly sample the depth probability volume to obtain the point confidence $\PointConf_\iP$ at each point $\PointX_\iP$. 
The above process can be expressed by
\begin{adjustwidth}{0pt}{0pt}
    \setlength{\abovedisplayskip}{-5pt}%
    \setlength{\abovedisplayshortskip}{\abovedisplayskip}%
    \begin{equation}
    \{\PointX_\iP,\PointConf_\iP\} = \NetPointPC(\Img_\iI, \Cam_\iI, \Img_{\iI_1}, \Cam_{\iI_1}, \Img_{\iI_2}, \Cam_{\iI_2},...),
    \end{equation}
\end{adjustwidth}
where $\NetPointPC$ is the MVSNet-based network. $\Img_{\iI_1}, \Cam_{\iI_1},...$ are additional neighboring views used in the MVS reconstruction; we use two additional views in most cases.

\boldstartspace{Point features.}
We use a 2D CNN $\NetPointF$ to extract neural 2D image feature maps from each image $\Img_\iI$.
These feature maps are aligned with the point (depth) prediction from $\NetPointPC$ and are used to directly predict per-point features $\PointF_\iP$ as:
 \begin{adjustwidth}{0pt}{0pt}
    \setlength{\abovedisplayskip}{-5pt}%
    \setlength{\abovedisplayshortskip}{\abovedisplayskip}%
    \setlength{\belowdisplayskip}{-5pt}%
    \begin{equation}
    \{\PointF_\iP\} = \NetPointF(\Img_\iI).
    \end{equation}
\end{adjustwidth}
In particular, we use a VGG network architecture for $\NetPointF$ that has three downsampling layers. We combine intermediate features at different resolutions as $\PointF_\iP$, providing a meaningful point description that models multi-scale scene appearance. (See Fig. \ref{fig:overview}(a))

\boldstartspace{End-to-end reconstruction.}
We combine point clouds from multiple viewpoints to obtain our final neural point cloud. We train the point generation networks along with the representation networks, from end to end with a rendering loss (see Fig. \ref{fig:pipeline}).
This allows our generation modules to produce reasonable initial radiance fields.
It also initializes the MLPs in our Point-NeRF representation with reasonable weights, significantly saving the per-scene fitting time. 

Moreover, apart from using the full generation module, our pipeline also supports using a point cloud reconstructed from other approaches like COLMAP \cite{schoenberger2016mvs}, where our model (excluding the MVS network) can still provide meaningful initial neural features for each point. Please refer to our supplementary material for the details.

\subsection{Optimizing point-based radiance fields}

\label{sec:prune_grow}
The above pipeline can output a reasonable initial point-based radiance field for a novel scene. Through differentiable ray marching, we can further improve the radiance field by optimizing the neural point cloud (point features $f_i$ and point confidence $\gamma_i$) and the MLPs in our representation, for that specific scene (see Fig. \ref{fig:pipeline}).

The initial point cloud, especially ones from external reconstruction methods (e.g., Metashape or COLMAP in Fig.~\ref{fig:teaser}), can often contain holes and outliers that degrade the rendering quality. 
During per-scene optimization, to solve this problem, we find that directly optimizing the location of the existing points makes the training unstable and cannot fill the large holes (see \ref{fig:teaser}). Instead, we apply novel point pruning and growing techniques that gradually improve both geometry modeling and rendering quality.

\boldstartspace{Point pruning.}
As introduced in Sec.~\ref{sec:pcl_render}, we designed point confidence values $\PointConf_\iP$ that describe whether a neural point is near a scene surface.
We utilize these confidence values to prune unnecessary outlier points.
Note that the point confidence is directly related to the per-point contribution in volume density regression (Eqn.~\ref{eqn:weighted_density}); as a result, low confidence reflects low volume density in a point's local region indicating that it is empty. Therefore, we prune points that have $\PointConf_\iP < 0.1$ every 10K iterations.

We also impose a sparsity loss on point confidence \cite{lombardi2019neural}: 
\begin{adjustwidth}{0pt}{0pt}
    \setlength{\abovedisplayskip}{-10pt}%
    \setlength{\abovedisplayshortskip}{\abovedisplayskip}%
    \setlength{\belowdisplayshortskip}{-15pt}%
    \begin{equation}
    \mathcal{L}_\text{sparse} = \cfrac{1}{|\gamma|}\sum_{\gamma_i}{[log(\gamma_i) + log(1-\gamma_i)}]
    \end{equation}
\end{adjustwidth}
which forces the confidence value to be close to either zero or one. 
As shown in Fig. \ref{fig:pg_hotdog}, this pruning technique can remove outlier points and reduce the corresponding artifacts.

\boldstartspace{Point growing.}
We also propose a novel technique to grow new points to cover missing scene geometry in the original point cloud.
Unlike point pruning that directly utilizes information from existing points, growing points requires recovering information in empty regions where no point exists.
We achieve this by progressively growing points near the point cloud boundary based on the local scene geometry modeled by our Point-NeRF representation. 

In particular, we leverage the per-ray shading locations ($\ShadX_\iS$ in Eqn.~\ref{eq:raymarching}) sampled in the ray marching to identify new point candidates. 
Specifically, we identify the shading location $\ShadX_{\iS_g}$ with the highest opacity along the ray:
 \begin{adjustwidth}{0pt}{0pt}
        \setlength{\abovedisplayskip}{-5pt}%
        \setlength{\abovedisplayshortskip}{\abovedisplayskip}%
        \setlength{\belowdisplayshortskip}{-10pt}%
    \begin{equation}
    \Contr_{\iS} = 1-\exp (-\Dens_\iS \Step_\iS), \;\; \iS_g = \argmax_\iS \Contr_{\iS}.
    \end{equation}
\end{adjustwidth}
We compute $\Dist_{\iS_g}$ as $\ShadX_{\iS_g}$'s distance to its closest neural point.

For a marching ray, we grow a neural point at $\ShadX_{\iS_g}$ if  $\Contr_{\iS_g}>\ThresContr$ and $\Dist_{\iS_g}>\ThresDist$. This implies that the location lies near the surface, but is far from other neural points. By repeating this growing strategy, our radiance field can be expanded to cover missing regions in the initial point cloud.
Point growing especially benefits point clouds reconstructed by methods like COLMAP that are not dense (see Fig. \ref{fig:pg_hotdog}). We show that even on an extreme case with only 1000 initial points, our technique is able to progressively grow new points and reasonably cover the object surface (see Fig. \ref{fig:chair_grow}).

\section{Implementation details}
\boldstart{Network details.}
We apply frequency positional encoding on the relative position and the per-point features for the per-point processing network $\NetPointF$, and the viewing direction for the network $\NetRad$.
We extract multi-scale images features from three layers at different resolutions in network $\NetPointF$, leading to a vector with 56 (8+16+32) channels. We additionally append the corresponding viewing directions from each input viewpoint, to handle view-dependent effects. Therefore our final per-point neural feature is a 59-channel vector. Please refer to our supplemental material for the details of network architectures and neural point querying during shading.

\boldstartspace{Training and optimization details.}
We train our full pipeline on the DTU dataset, using the same training and testing split as PixelNeRF and MVSNeRF.
We first pretrain the MVSNet-based depth generation network using the ground truth depth similar to the original MVSNet paper \cite{yao2018mvsnet}. We then train our full pipeline from end to end purely with a L2 rendering loss $\mathcal{L}_\text{render}$, supervising our rendered pixels from ray marching (via Eqn.~\ref{eq:raymarching}) with the ground truth, to obtain our Point-NeRF reconstruction network.
We train our full pipeline using Adam \cite{kingma2014adam} optimizer with an initial learning rate of $5e^{-4}$. Our feed-forward network takes $0.2s$ to generate a point cloud from three input views.

In the per-scene optimization stage, we adopt a loss function that combines the rendering and the sparsity loss
\begin{adjustwidth}{0pt}{0pt}
    \setlength{\abovedisplayskip}{-10pt}%
    \setlength{\abovedisplayshortskip}{\abovedisplayskip}%
    \setlength{\belowdisplayshortskip}{-10pt}%
    \begin{equation}
    \mathcal{L}_\text{opt} = \mathcal{L}_\text{render} + a \mathcal{L}_\text{sparse},
    \end{equation}
\end{adjustwidth}
where we use $a=2e^{-3}$ ~for all our experiments. We perform point growing and pruning every 10K iterations to achieve our final high-quality reconstruction.
    \begin{table*} [ht]
        \begin{adjustwidth}{0pt}{0pt}  
        \centering
        \captionsetup{aboveskip=5pt}
            \setlength\tabcolsep{4pt}
            {\small
            \begin{tabular}{l|cccc|ccccc}
            \hline
                  & \multicolumn{4}{c|}{No Per-scene Optimization} & \multicolumn{5}{c}{Per-scene Optimization}                    \\
                  & PixelNeRF\cite{yu2020pixelnerf}   &  MVSNeRF\cite{chen2021mvsnerf} & IBRNet \cite{ibrnet} & Ours  & Ours$_{1K}$ & Ours$_{10K}$  & MVSNeRF$_{10K}$ & IBRNet$_{10K}$ & NeRF$_{200k}$  \\ \hline
            PSNR $\uparrow$  & 19.31       & \bone 26.63    & \btwo 26.04   & \bthird 23.89       & 28.43     & \btwo 30.12    & \bthird 28.50           & \bone \textbf{31.35}     & 27.01 \\
            SSIM $\uparrow$  & 0.789       & \bone 0.931    & \btwo 0.917   & \bthird 0.874       & 0.929         & \bone \textbf{0.957}  & \bthird 0.933    & \btwo 0.956      & 0.902 \\
            LPIPS$_{Vgg}$ $\downarrow$ & 0.382       & \bone 0.168    & \btwo 0.190    & \bthird 0.203       & 0.183     & \bone \textbf{0.117}    & \bthird 0.179           & \btwo 0.131      & 0.263 \\
            Time$ \downarrow$   & -           & -        & -       & -           & \bone \textbf{2min}  & \bthird 20min   &  \btwo 24min       & 1h         & 10h   \\ \hline
            \end{tabular} 
            }
            \captionsetup{aboveskip=2pt}
            \captionsetup{belowskip=-5pt}
            \caption {Comparisons of our Point-NeRF with radiance-based models \cite{martin2021nerf,ibrnet,liu2020neural} and a point-based rendering model \cite{aliev2020neural} on the DTU dataset \cite{dtu} with the novel view synthesis setting introduced in \cite{chen2021mvsnerf}. The subscripts indicate the number of iterations during optimization.} 
            \label{tb:dtu}
        \end{adjustwidth}
    \end{table*}
    
    \begin{table*}[ht]
        \begin{adjustwidth}{0pt}{0pt}  
        \centering
            \small{
                \begin{tabular}{l|ccccccc}
                \hline 
                & NPBG\cite{aliev2020neural} & NeRF \cite{martin2021nerf} & IBRNet \cite{ibrnet} & NSVF \cite{liu2020neural}   & Point-NeRF$^{col}_{200K}$ & Point-NeRF$_{20K}$ & Point-NeRF$_{200K}$ \\ \hline
                PSNR $\uparrow$   & 24.56 & \bthird 31.01 & 28.14 & \btwo 31.75  & 31.77 &  30.71  & \bone \textbf{33.31} \\
                SSIM $\uparrow$       & 0.923 & 0.947 & 0.942 & \bthird \bthird 0.964 & \btwo 0.973 &   0.967  & \bone \textbf{0.978} \\
                LPIPS$_{Vgg}$ $\downarrow$ & 0.109 & 0.081 & \bthird 0.072  & - &  \btwo 0.062    &  0.081  & \bone \textbf{0.049} \\
                LPIPS$_{Alex}$ $\downarrow$ & 0.095 & - & -  & \bthird 0.047 & \btwo 0.040 &   0.050 & \bone \textbf{0.027} \\ \hline
                \end{tabular}
            }
             \captionsetup{aboveskip=3pt}
            \captionsetup{belowskip=-10pt}
            \caption {Comparisons of Point-NeRF with radiance-based models \cite{martin2021nerf,ibrnet,liu2020neural} and a point-based rendering model \cite{aliev2020neural} on the Synthetic-NeRF dataset \cite{martin2021nerf}. The subscripts indicate the number of iterations. Our model not only surpasses other methods when converged after $200K$ steps (Point-NeRF$_{200K}$), but surpasses IBRNet \cite{ibrnet} and is on par with NeRF \cite{mildenhall2020nerf} when optimized by only $20K$ steps (Point-NeRF$_{20K}$). Our methods can also initialize radiance fields based on point clouds reconstructed by methods such as COLMAP (Point-NeRF$^{col}_{200K}$).}
            \label{tb:nerfsynth} 
        \end{adjustwidth}
    \end{table*}
    \section{Experiments}
    \subsection{Evaluation on the DTU testing set.}
    We evaluate our model on the DTU testing set. We produce novel view synthesis from both direct network inference and per-scene fine-tuning optimization, and compare them with the previous state-of-the art methods including PixelNeRF\cite{yu2020pixelnerf}, IBRNet\cite{ibrnet}, MVSNeRF\cite{chen2021mvsnerf}, and NeRF\cite{mildenhall2020nerf}. 
    IBRNet and MVSNeRF utilize similar per-scene fine-tuning; we fine-tune all methods with 10k iterations for the comparison. Additionally, we show our results with only 1k iterations to demonstrate the optimization efficiency.
    
    Tab.~\ref{tb:dtu} shows the quantitative results of all methods with PSNR, SSIM, and LPIPS; qualitative rendering results are shown in Fig.~\ref{fig:dtu}.
    We can see that our fine-tuning results after 10k iterations achieve the best SSIM and LPIPS\cite{zhang2018perceptual}, two out of the three metrics. These are significantly better than MVSNeRF and NeRF.
    While IBRNet produces slightly better PSNRs, our final renderings in fact recover more accurate texture details and highlights as shown Fig.~\ref{fig:dtu}.
    On the other hand, IBRNet is also more expensive to fine-tune, taking 1 hour---5x longer than ours for the same iterations. This is because IBRNet utilizes a large global CNN, whereas Point-NeRF leverages local point features with small MLPs that are easier to optimize. More importantly, our neural points lies near actual scene surfaces, thus avoids sampling ray points in the empty space.
    
    Apart from the optimization results, our initial radiance field estimated from our network is significantly better than PixelNeRF. In this case, our direct inference is worse than IBRNet and MVSNet, mainly because these two methods are using more complex variance-based feature extraction. Our point features are extracted from a simple VGG network. The same design is used in PixelNeRF; we achieve significantly better results than PixelNeRF due to our novel surface-adaptive point-based representation.
    
    While a more complex feature extractor as in IBRNet might improve quality, it will add burden to memory usage and training efficiency. 
    More importantly, our generation network has already provided high-quality initial radiance field to support efficient optimization. We show that with even 2 min / 1K iterations of fine-tuning for our method leading to a very high visual quality comparable to MVSNeRF's final 10k-iteration results. This clearly demonstrates the high reconstruction efficiency of our approach.
    
    \subsection{Evaluation on the NeRF Synthetic dataset.} 
    While our model is purely trained on the DTU
    dataset, our network generalizes well to novel datasets that have completely different camera distributions. 
    We demonstrate such results on the NeRF synthetic dataset and compare with other methods with qualitative results in Fig.~\ref{fig:nerfsynth} and quantitative results in Tab.~\ref{tb:nerfsynth}.
    We compare with a point-based rendering model (NPBG) \cite{aliev2020neural}, a generalizable radiance field method (IBRNet) \cite{ibrnet}, and per-scene radiance field reconstruction techniques (NeRF and NSVF)\cite{mildenhall2020nerf,liu2020neural}.
    
    \boldstartspace{Comparisons with generalizing methods.}
    We compare with IBRNet, to the best of our knowledge, is the previous best NeRF-based generalizable model that can handle free-viewpoint rendering with any arbitrary numbers. 
    Note that, this dataset has a $360^{\circ}$ camera distribution, which is much wider than the DTU dataset. 
    In this case, methods like MVSNeRF cannot be applied, since it recovers a local perspective frustum volume from three input images, which cannot cover the entire $360^{\circ}$ viewing range.
    We, therefore, compare with IBRNet and focus on final results after per-scene optimization in this experiment. We use their released model to produce the results. 
    Our results at 20k iterations (Point-NeRF$_{20K}$) have already outperformed IBRNet's converged results with better PSNR, SSIM, and LIPIPS; we also achieve rendering quality with better geometry and texture details as shown in Fig.~\ref{fig:nerfsynth}. 
    
    \boldstartspace{Comparisons with pure per-scene methods.}
    Our results after 20K iterations are quantitatively very close to NeRF's results trained with 200K iterations. Visually, our model at $20K$ iterations already has better renderings in some cases, e.g. the Ficus scene (4th row) in Fig.~\ref{fig:nerfsynth}. Point-NeRF$_{20K}$ is optimized for only 40 minutes, which is at least $30\times$ faster than the 20+ hours optimization time taken by NeRF.
    NSVF's~\cite{liu2020neural} results are also from very long per-scene optimization and yet are only slightly better than our 40min results.
    Optimizing our model for 200K until convergence can lead to significantly better results than NeRF, NSVF, and all other comparison methods. As shown in Fig.~\ref{fig:nerfsynth}, our 200K results contain the most geometry and texture details. Attribute to the point growing technique, our method is the only one that can fully recover details like the thin rope structure in the Ship scene (2nd row). 
    
    \boldstartspace{Comparisons with point-based rendering.}
    Our results are significantly better than the previous state-of-the-art point-based rendering methods. We run NPBG\cite{aliev2020neural} using the same point cloud generated by our MVSNet-based network. However, NPBG can only produce blurry rendering results with their rasterization and 2D CNN framework.
    In contrast, we leverage volumetric rendering technique with neural radiance fields, leading to photo-realistic results.
    
    \subsection{Evaluation on the Tanks \& Temples and the ScanNet dataset.}
    We compare Point-NeRF with NSVF on the Tanks \& Temples and the ScanNet dataset in Tab. \ref{tb:ttsn}. Please refer to the supplemental materials for more comparisons.
    \begin{table}[ht]
        \begin{adjustwidth}{0pt}{0pt}  
        \centering
        \setlength\tabcolsep{4pt}
        {\small
            \begin{tabular}{l|c|c}
            \hline
             & Tanks \& Temples\cite{Knapitsch2017}  &  ScanNet\cite{dai2017scannet}     \\ \hline
             NSVF\cite{liu2020neural}  & 28.40 / 0.900 / 0.153  & 25.48 / 0.688 / 0.301 \\ \hline
             Point-NeRF & \textbf{29.61} / \textbf{0.954} / \textbf{0.080} & \textbf{30.32} / \textbf{0.909} / \textbf{0.220} \\ \hline
            \end{tabular}
        }
        \captionsetup{aboveskip=2pt}
        \captionsetup{belowskip=-10pt}
        \caption {The quantitative results (PSNR / SSIM / LPIPS$_{Alex}$) on the Tanks \& Temples and the ScanNet dataset.}
        \label{tb:ttsn} 
        \end{adjustwidth}
    \end{table}
\begin{figure}
        \vspace{-5pt}
        \begin{adjustwidth}{-0pt}{0pt}
            \begin{center}
                \includegraphics[width=1.\linewidth]{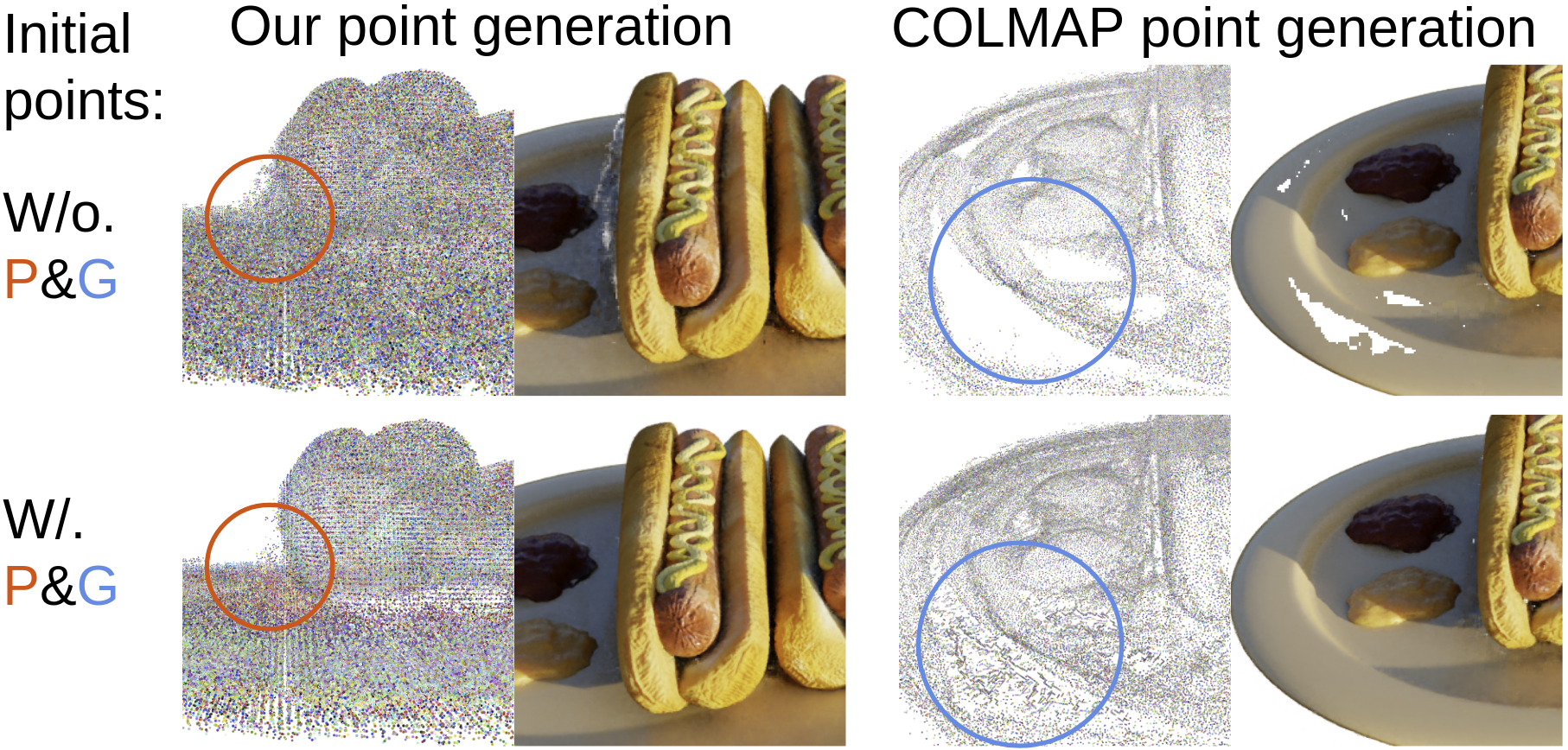}
            \end{center}
        \end{adjustwidth}
        \captionsetup{aboveskip=5pt}
        \captionsetup{belowskip=-2pt}
        \caption{Our neural point clouds and rendered novel views with or without point pruning and growing (P\&G). P\&G improves both the geometries and rendering results when using the point cloud reconstructed from our model or from COLMAP\cite{schoenberger2016mvs}.}
        \label{fig:pg_hotdog}
    \end{figure}

\subsection{Additional experiments.}

    \boldstartspace{Converting COLMAP point clouds to Point-NeRF}
    Apart from using our full pipeline, Point-NeRF can also be used to convert standard point clouds reconstructed by other techniques to point-based radiance fields. We run experiments for this on the full NeRF synthetic dataset, using the point cloud reconstructed by COLMAP \cite{schoenberger2016mvs}. 
    The quantitative results are shown as Point-NeRF$_{col}$ in Tab.~\ref{tb:nerfsynth}.
    Since COLMAP point clouds may contain a lot of holes (as shown in Fig.~\ref{fig:teaser}) and noises, we optimize the model for 200K after the initialization to address the point cloud issues with our point growing and pruning techniques.
    Note that, even from this low-quality point cloud, our final results are still of very high quality with very high SSIM and LPIPS numbers compared to all other methods. This demonstrates that our technique can be potentially combined with any existing point cloud reconstruction techniques, to achieve realistic rendering while improving the point cloud geometry.
    
    \begin{figure}
        \vspace{-5pt}
        \begin{adjustwidth}{-0pt}{0pt}
            \begin{center}
                \includegraphics[width=1.\linewidth]{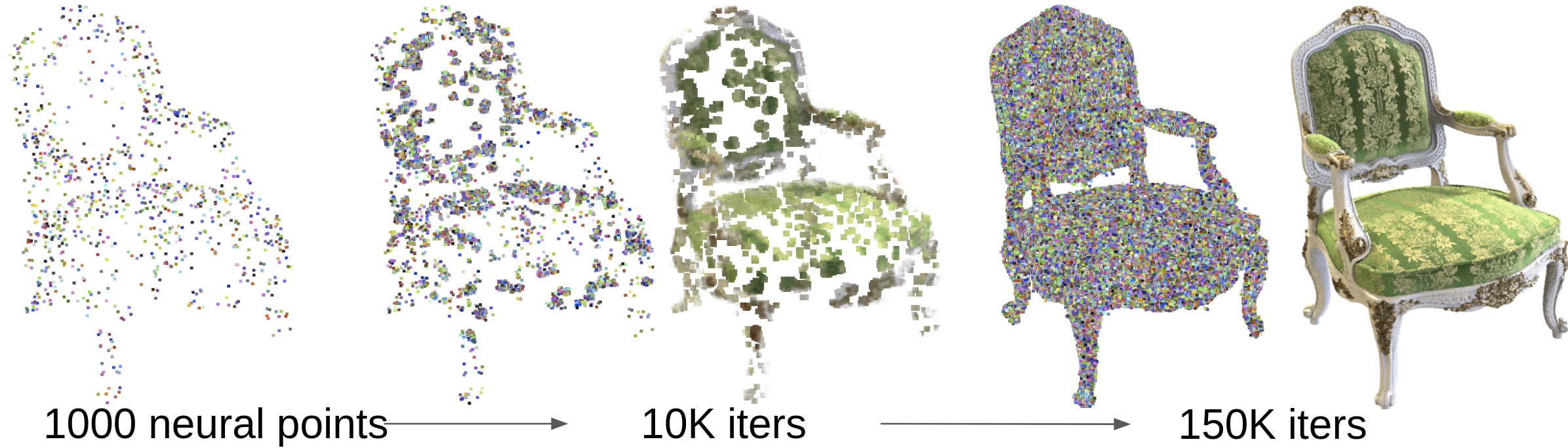}
            \end{center}
        \end{adjustwidth}
        \captionsetup{aboveskip=5pt}
        \captionsetup{belowskip=-15pt}
        \caption{Starting from 1000 randomly sampled COLMAP points of the Chair scene, our point growing mechanism can help complete the geometry and generate high-quality novel views when only being supervised by RGB images.}
        \label{fig:chair_grow}
    \end{figure}
    
    \boldstartspace{Point growing and pruning.}
    To further demonstrate the effectiveness of our point growing and pruning modules, we show ablation study results with and without the point growing and pruning in the per-scene optimization. We conduct this experiment on the Hotdog and Ship scenes, using both our full model and our model with COLMAP point clouds. The quantitative results are shown in Tab.~\ref{tb:ab_pg}; our point growing and pruning techniques are very effective, significantly improving the reconstruction results on both cases. We also show the visual results of the Hotdog scene in Fig.~\ref{fig:pg_hotdog}. We can clearly see that our model is able to prune the point outliers on the left and successfully fill the severe holes on the right in the original COLMAP point cloud.
    
    We also manually create an extreme example to show our point growing technique in Fig.~\ref{fig:chair_grow}, where we start from a very sparse point cloud with only 1000 points sampled from our original point reconstruction. We demonstrate that our approach can progressively grow new point from the point cloud boundary until filling the entire scene surface through iterations. This example further demonstrates the effectiveness of our model, which has high potentials in using image data to recover the accurate scene geometry and appearance from low-quality point clouds. %
    
    Please find more results in the supplemental materials.
    
    \begin{figure*}
        \vspace{-10pt}
        \begin{adjustwidth}{-0pt}{0pt}
            \begin{center}
                \includegraphics[width=0.93\textwidth]{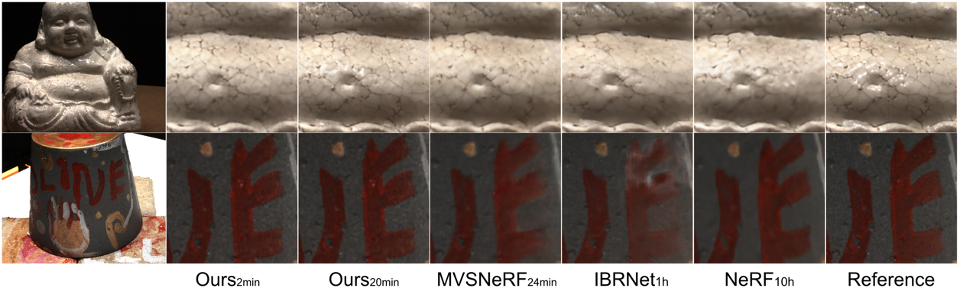}
            \end{center}
        \end{adjustwidth}
        \captionsetup{aboveskip=3pt}
        \captionsetup{belowskip=-2pt}
        \caption{Qualitative comparisons of per-scene optimization on the DTU dataset \cite{dtu}. Our Point-NeRF can recover texture details and geometrical structures more accurately than other methods. Point-NeRF also demonstrates superior efficiency. Within two mins, our model trained for 1K steps is already on par with the state-of-the-art methods such as MVSNeRF \cite{chen2021mvsnerf} and IBRNet\cite{ibrnet}} 
        \label{fig:dtu}
    \end{figure*}

    \begin{figure*}
        \begin{adjustwidth}{-0pt}{0pt}
            \begin{center}
                \includegraphics[width=0.93\textwidth]{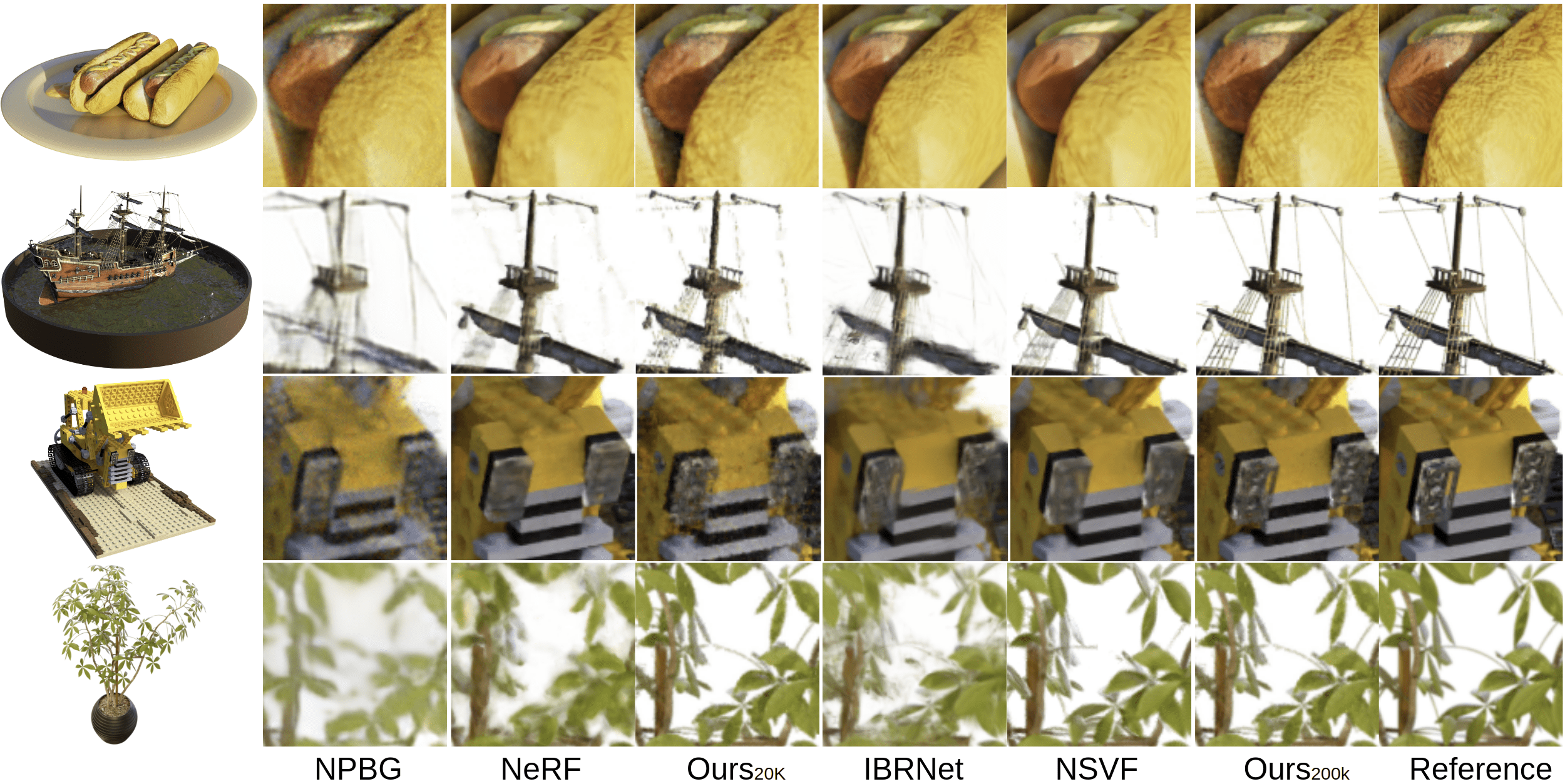}
            \end{center}
        \end{adjustwidth}
         \captionsetup{aboveskip=3pt}
        \captionsetup{belowskip=-7pt}
        \caption{Qualitative comparisons on the NeRF Synthetic dataset \cite{mildenhall2020nerf}. The subscripts indicate the number of iterations. Our Point-NeRF can capture fine details and thin structures (see the rope on row 2). Point-NeRF also demonstrates superior efficiency. Our model trained for 20K steps already on par with NeRF with $30\times$ faster training time.}
        \label{fig:nerfsynth}
    \end{figure*}

     \begin{table}[ht]
        \begin{adjustwidth}{0pt}{0pt}  
        \centering
        \setlength\tabcolsep{4pt}
        {\small
            \begin{tabular}{lc|c|c}
            \hline
            Method & P\&G & Ship &  Hotdog      \\ \hline
            Ours        & No   & 25.50 / 0.878 / 0.182          & 34.91 / 0.983 / 0.067           \\
            Ours        & Yes  & \textbf{30.97} / \textbf{0.942} / \textbf{0.124} & \textbf{37.30} / \textbf{0.991} / \textbf{0.037} \\ \hline
            COLMAP      & No   & 19.35 / 0.905 / 0.167          & 29.91 / 0.978 / 0.061          \\
            COLMAP      & Yes  & \textbf{30.18} / \textbf{0.941} / \textbf{0.134} & \textbf{35.49} / \textbf{0.986} / \textbf{0.061} \\ \hline
            \end{tabular}
        }
        \captionsetup{aboveskip=2pt}
        \captionsetup{belowskip=-10pt}
        \caption {The quantitative results (PSNR / SSIM / LPIPS$_{Vgg}$) of the Ship and Hotdog scene with or without point pruning and growing (P\&G). The improvements are significant when using either our generated points or the point cloud generated by  COLMAP\cite{schoenberger2016mvs}.}
        \label{tb:ab_pg} 
        \end{adjustwidth}
    \end{table}

\section{Conclusion}
In this paper, we present a novel approach for high-quality neural scene reconstruction and rendering.
We propose a novel neural scene representation---Point-NeRF---that models a volumetric radiance field with a neural point cloud.
We reconstruct a good initialization of Point-NeRF directly from input images via direct network inference and show that we can efficiently finetune this initialization for a scene.
This enables highly efficient Point-NeRF reconstruction with only 20--40 min per-scene optimization, leading to rendering quality comparable to and even surpassing NeRF that requires substantially longer training time (20+ hours).
We also present novel effective growing and pruning techniques for our per-scene optimization, significantly improving our results and making our approach robust with different point cloud quality.
Our Point-NeRF successfully combines the advantages from both classical point cloud representation and neural radiance field representation, making an important step towards a practical scene reconstruction solution with high efficiency and realism. 


{\small
\bibliographystyle{ieee_fullname}
\bibliography{egbib}

\begin{thebibliography}{10}\itemsep=-1pt

\bibitem{achlioptas2018learning}
Panos Achlioptas, Olga Diamanti, Ioannis Mitliagkas, and Leonidas Guibas.
\newblock Learning representations and generative models for {3D} point clouds.
\newblock In {\em ICML}, pages 40--49, 2018.

\bibitem{aliev2020neural}
Kara-Ali Aliev, Artem Sevastopolsky, Maria Kolos, Dmitry Ulyanov, and Victor
  Lempitsky.
\newblock Neural point-based graphics.
\newblock In {\em Computer Vision--ECCV 2020: 16th European Conference,
  Glasgow, UK, August 23--28, 2020, Proceedings, Part XXII 16}, pages 696--712.
  Springer, 2020.

\bibitem{bi2020neural}
Sai Bi, Zexiang Xu, Pratul Srinivasan, Ben Mildenhall, Kalyan Sunkavalli,
  Milo{\v{s}} Ha{\v{s}}an, Yannick Hold-Geoffroy, David Kriegman, and Ravi
  Ramamoorthi.
\newblock Neural reflectance fields for appearance acquisition.
\newblock {\em arXiv preprint arXiv:2008.03824}, 2020.

\bibitem{bi2020deep}
Sai Bi, Zexiang Xu, Kalyan Sunkavalli, Milo{\v{s}} Ha{\v{s}}an, Yannick
  Hold-Geoffroy, David Kriegman, and Ravi Ramamoorthi.
\newblock Deep reflectance volumes: Relightable reconstructions from multi-view
  photometric images.
\newblock In {\em Proc.~ECCV}, 2020.

\bibitem{boss2021nerd}
Mark Boss, Raphael Braun, Varun Jampani, Jonathan~T Barron, Ce Liu, and Hendrik
  Lensch.
\newblock Nerd: Neural reflectance decomposition from image collections.
\newblock In {\em Proceedings of the IEEE/CVF International Conference on
  Computer Vision}, pages 12684--12694, 2021.

\bibitem{buehler2001unstructured}
Chris Buehler, Michael Bosse, Leonard McMillan, Steven Gortler, and Michael
  Cohen.
\newblock Unstructured lumigraph rendering.
\newblock In {\em Proc.~SIGGRAPH}, pages 425--432, 2001.

\bibitem{chan2021pi}
Eric~R Chan, Marco Monteiro, Petr Kellnhofer, Jiajun Wu, and Gordon Wetzstein.
\newblock pi-gan: Periodic implicit generative adversarial networks for
  3d-aware image synthesis.
\newblock In {\em Proceedings of the IEEE/CVF Conference on Computer Vision and
  Pattern Recognition}, pages 5799--5809, 2021.

\bibitem{chen2021mvsnerf}
Anpei Chen, Zexiang Xu, Fuqiang Zhao, Xiaoshuai Zhang, Fanbo Xiang, Jingyi Yu,
  and Hao Su.
\newblock Mvsnerf: Fast generalizable radiance field reconstruction from
  multi-view stereo.
\newblock {\em arXiv preprint arXiv:2103.15595}, 2021.

\bibitem{chen2018learning}
Zhiqin Chen and Hao Zhang.
\newblock Learning implicit fields for generative shape modeling.
\newblock In {\em Proc.~CVPR}, 2019.

\bibitem{cheng2020deep}
Shuo Cheng, Zexiang Xu, Shilin Zhu, Zhuwen Li, Li~Erran Li, Ravi Ramamoorthi,
  and Hao Su.
\newblock Deep stereo using adaptive thin volume representation with
  uncertainty awareness.
\newblock In {\em Proceedings of the IEEE/CVF Conference on Computer Vision and
  Pattern Recognition}, pages 2524--2534, 2020.

\bibitem{dai2017scannet}
Angela Dai, Angel~X Chang, Manolis Savva, Maciej Halber, Thomas Funkhouser, and
  Matthias Nie{\ss}ner.
\newblock Scannet: Richly-annotated 3d reconstructions of indoor scenes.
\newblock In {\em Proceedings of the IEEE conference on computer vision and
  pattern recognition}, pages 5828--5839, 2017.

\bibitem{debevec1998efficient}
Paul Debevec, Yizhou Yu, and George Borshukov.
\newblock Efficient view-dependent image-based rendering with projective
  texture-mapping.
\newblock In {\em Rendering Techniques’ 98}, pages 105--116. 1998.

\bibitem{drebin1988volume}
Robert~A Drebin, Loren Carpenter, and Pat Hanrahan.
\newblock Volume rendering.
\newblock {\em ACM Siggraph Computer Graphics}, 22(4):65--74, 1988.

\bibitem{furukawa2009accurate}
Yasutaka Furukawa and Jean Ponce.
\newblock Accurate, dense, and robust multiview stereopsis.
\newblock {\em IEEE Transactions on Pattern Analysis and Machine Intelligence},
  32(8):1362--1376, 2009.

\bibitem{he2015delving}
Kaiming He, Xiangyu Zhang, Shaoqing Ren, and Jian Sun.
\newblock Delving deep into rectifiers: Surpassing human-level performance on
  imagenet classification.
\newblock In {\em Proceedings of the IEEE international conference on computer
  vision}, pages 1026--1034, 2015.

\bibitem{hedman2021baking}
Peter Hedman, Pratul~P Srinivasan, Ben Mildenhall, Jonathan~T Barron, and Paul
  Debevec.
\newblock Baking neural radiance fields for real-time view synthesis.
\newblock {\em arXiv preprint arXiv:2103.14645}, 2021.

\bibitem{huang2018deepmvs}
Po-Han Huang, Kevin Matzen, Johannes Kopf, Narendra Ahuja, and Jia-Bin Huang.
\newblock Deepmvs: Learning multi-view stereopsis.
\newblock In {\em Proceedings of the IEEE Conference on Computer Vision and
  Pattern Recognition}, pages 2821--2830, 2018.

\bibitem{dtu}
Rasmus Jensen, Anders Dahl, George Vogiatzis, Engil Tola, and Henrik Aan{\ae}s.
\newblock Large scale multi-view stereopsis evaluation.
\newblock In {\em 2014 CVPR}, pages 406--413. IEEE, 2014.

\bibitem{ji2017surfacenet}
Mengqi Ji, Juergen Gall, Haitian Zheng, Yebin Liu, and Lu Fang.
\newblock {SurfaceNet}: An end-to-end {3D} neural network for multiview
  stereopsis.
\newblock In {\em Proc.~ICCV}, 2017.

\bibitem{kanazawa2018learning}
Angjoo Kanazawa, Shubham Tulsiani, Alexei~A Efros, and Jitendra Malik.
\newblock Learning category-specific mesh reconstruction from image
  collections.
\newblock In {\em Proc.~ECCV}, 2018.

\bibitem{kazhdan2006poisson}
Michael Kazhdan, Matthew Bolitho, and Hugues Hoppe.
\newblock Poisson surface reconstruction.
\newblock In {\em Proc. Eurographics Symposium on Geometry Processing},
  volume~7, 2006.

\bibitem{kingma2014adam}
Diederik~P Kingma and Jimmy Ba.
\newblock Adam: A method for stochastic optimization.
\newblock {\em arXiv preprint arXiv:1412.6980}, 2014.

\bibitem{Knapitsch2017}
Arno Knapitsch, Jaesik Park, Qian-Yi Zhou, and Vladlen Koltun.
\newblock Tanks and temples: Benchmarking large-scale scene reconstruction.
\newblock {\em ACM Transactions on Graphics}, 36(4), 2017.

\bibitem{kopanas2021point}
Georgios Kopanas, Julien Philip, Thomas Leimk{\"u}hler, and George Drettakis.
\newblock Point-based neural rendering with per-view optimization.
\newblock In {\em Computer Graphics Forum}, volume~40, pages 29--43. Wiley
  Online Library, 2021.

\bibitem{kutulakos2000theory}
Kiriakos~N Kutulakos and Steven~M Seitz.
\newblock A theory of shape by space carving.
\newblock {\em International Journal of Computer Vision}, 38(3):199--218, 2000.

\bibitem{lassner2021pulsar}
Christoph Lassner and Michael Zollhofer.
\newblock Pulsar: Efficient sphere-based neural rendering.
\newblock In {\em Proceedings of the IEEE/CVF Conference on Computer Vision and
  Pattern Recognition}, pages 1440--1449, 2021.

\bibitem{li2021neural}
Zhengqi Li, Simon Niklaus, Noah Snavely, and Oliver Wang.
\newblock Neural scene flow fields for space-time view synthesis of dynamic
  scenes.
\newblock In {\em Proceedings of the IEEE/CVF Conference on Computer Vision and
  Pattern Recognition}, pages 6498--6508, 2021.

\bibitem{liu2015learning}
Fayao Liu, Chunhua Shen, Guosheng Lin, and Ian Reid.
\newblock Learning depth from single monocular images using deep convolutional
  neural fields.
\newblock {\em IEEE Transactions on Pattern Analysis and Machine Intelligence},
  38(10):2024--2039, 2016.

\bibitem{liu2020neural}
Lingjie Liu, Jiatao Gu, Kyaw~Zaw Lin, Tat-Seng Chua, and Christian Theobalt.
\newblock Neural sparse voxel fields.
\newblock {\em arXiv preprint arXiv:2007.11571}, 2020.

\bibitem{lombardi2019neural}
Stephen Lombardi, Tomas Simon, Jason Saragih, Gabriel Schwartz, Andreas
  Lehrmann, and Yaser Sheikh.
\newblock Neural volumes: Learning dynamic renderable volumes from images.
\newblock {\em arXiv preprint arXiv:1906.07751}, 2019.

\bibitem{lorensen1987marching}
William~E Lorensen and Harvey~E Cline.
\newblock Marching cubes: A high resolution 3d surface construction algorithm.
\newblock {\em SIGGRAPH Computer Graphics}, 21(4):163--169, 1987.

\bibitem{martin2021nerf}
Ricardo Martin-Brualla, Noha Radwan, Mehdi~SM Sajjadi, Jonathan~T Barron,
  Alexey Dosovitskiy, and Daniel Duckworth.
\newblock Nerf in the wild: Neural radiance fields for unconstrained photo
  collections.
\newblock In {\em Proceedings of the IEEE/CVF Conference on Computer Vision and
  Pattern Recognition}, pages 7210--7219, 2021.

\bibitem{mescheder2018occupancy}
Lars Mescheder, Michael Oechsle, Michael Niemeyer, Sebastian Nowozin, and
  Andreas Geiger.
\newblock Occupancy networks: Learning 3d reconstruction in function space.
\newblock {\em Proc.~CVPR}, 2019.

\bibitem{meshry2019neural}
Moustafa Meshry, Dan~B Goldman, Sameh Khamis, Hugues Hoppe, Rohit Pandey, Noah
  Snavely, and Ricardo Martin-Brualla.
\newblock Neural rerendering in the wild.
\newblock In {\em Proceedings of the IEEE/CVF Conference on Computer Vision and
  Pattern Recognition}, pages 6878--6887, 2019.

\bibitem{mildenhall2020nerf}
Ben Mildenhall, Pratul~P Srinivasan, Matthew Tancik, Jonathan~T Barron, Ravi
  Ramamoorthi, and Ren Ng.
\newblock Nerf: Representing scenes as neural radiance fields for view
  synthesis.
\newblock In {\em European conference on computer vision}, pages 405--421.
  Springer, 2020.

\bibitem{niemeyer2021giraffe}
Michael Niemeyer and Andreas Geiger.
\newblock Giraffe: Representing scenes as compositional generative neural
  feature fields.
\newblock In {\em Proceedings of the IEEE/CVF Conference on Computer Vision and
  Pattern Recognition}, pages 11453--11464, 2021.

\bibitem{niemeyer2020differentiable}
Michael Niemeyer, Lars Mescheder, Michael Oechsle, and Andreas Geiger.
\newblock Differentiable volumetric rendering: Learning implicit 3d
  representations without 3d supervision.
\newblock In {\em Proc.~CVPR}, 2020.

\bibitem{park2021nerfies}
Keunhong Park, Utkarsh Sinha, Jonathan~T Barron, Sofien Bouaziz, Dan~B Goldman,
  Steven~M Seitz, and Ricardo Martin-Brualla.
\newblock Nerfies: Deformable neural radiance fields.
\newblock In {\em Proceedings of the IEEE/CVF International Conference on
  Computer Vision}, pages 5865--5874, 2021.

\bibitem{park2021hypernerf}
Keunhong Park, Utkarsh Sinha, Peter Hedman, Jonathan~T Barron, Sofien Bouaziz,
  Dan~B Goldman, Ricardo Martin-Brualla, and Steven~M Seitz.
\newblock Hypernerf: A higher-dimensional representation for topologically
  varying neural radiance fields.
\newblock {\em arXiv preprint arXiv:2106.13228}, 2021.

\bibitem{qi2017pointnet}
Charles~R Qi, Hao Su, Kaichun Mo, and Leonidas~J Guibas.
\newblock Pointnet: Deep learning on point sets for 3d classification and
  segmentation.
\newblock In {\em Proc.~CVPR}, 2017.

\bibitem{qi2016volumetric}
Charles~R Qi, Hao Su, Matthias Nie{\ss}ner, Angela Dai, Mengyuan Yan, and
  Leonidas~J Guibas.
\newblock Volumetric and multi-view cnns for object classification on 3d data.
\newblock In {\em Proc.~CVPR}, 2016.

\bibitem{reiser2021kilonerf}
Christian Reiser, Songyou Peng, Yiyi Liao, and Andreas Geiger.
\newblock Kilonerf: Speeding up neural radiance fields with thousands of tiny
  mlps.
\newblock {\em arXiv preprint arXiv:2103.13744}, 2021.

\bibitem{schoenberger2016sfm}
Johannes~Lutz Sch\"{o}nberger and Jan-Michael Frahm.
\newblock Structure-from-motion revisited.
\newblock In {\em Proc.~CVPR}, 2016.

\bibitem{schoenberger2016mvs}
Johannes~Lutz Sch\"{o}nberger, Enliang Zheng, Marc Pollefeys, and Jan-Michael
  Frahm.
\newblock {Pixelwise View Selection for Unstructured Multi-View Stereo}.
\newblock In {\em European Conference on Computer Vision (ECCV)}, 2016.

\bibitem{schwarz2020graf}
Katja Schwarz, Yiyi Liao, Michael Niemeyer, and Andreas Geiger.
\newblock Graf: Generative radiance fields for 3d-aware image synthesis.
\newblock {\em arXiv preprint arXiv:2007.02442}, 2020.

\bibitem{seitz1999photorealistic}
Steven~M Seitz and Charles~R Dyer.
\newblock Photorealistic scene reconstruction by voxel coloring.
\newblock {\em International Journal of Computer Vision}, 35(2):151--173, 1999.

\bibitem{sitzmann2019deepvoxels}
Vincent Sitzmann, Justus Thies, Felix Heide, Matthias Nie{\ss}ner, Gordon
  Wetzstein, and Michael Zollhofer.
\newblock Deepvoxels: Learning persistent {3D} feature embeddings.
\newblock In {\em Proc.~CVPR}, 2019.

\bibitem{sitzmann2019scene}
Vincent Sitzmann, Michael Zollh{\"o}fer, and Gordon Wetzstein.
\newblock Scene representation networks: Continuous 3d-structure-aware neural
  scene representations.
\newblock {\em arXiv preprint arXiv:1906.01618}, 2019.

\bibitem{tang2018ba}
Chengzhou Tang and Ping Tan.
\newblock {BA}-net: Dense bundle adjustment network.
\newblock In {\em Proc.~ICLR}, 2019.

\bibitem{vijayanarasimhan2017sfm}
Sudheendra Vijayanarasimhan, Susanna Ricco, Cordelia Schmid, Rahul Sukthankar,
  and Katerina Fragkiadaki.
\newblock Sfm-net: Learning of structure and motion from video.
\newblock {\em arXiv preprint arXiv:1704.07804}, 2017.

\bibitem{wang2018mvpnet}
Jinglu Wang, Bo Sun, and Yan Lu.
\newblock {MVP}net: Multi-view point regression networks for {3D} object
  reconstruction from a single image.
\newblock {\em Proc. AAAI Conference on Artificial Intelligence}, 2019.

\bibitem{wang2018pixel2mesh}
Nanyang Wang, Yinda Zhang, Zhuwen Li, Yanwei Fu, Wei Liu, and Yu-Gang Jiang.
\newblock Pixel2mesh: Generating 3d mesh models from single {RGB} images.
\newblock In {\em Proc.~ECCV}, 2018.

\bibitem{ibrnet}
Qianqian Wang, Zhicheng Wang, Kyle Genova, Pratul Srinivasan, Howard Zhou,
  Jonathan~T. Barron, Ricardo Martin-Brualla, Noah Snavely, and Thomas
  Funkhouser.
\newblock Ibrnet: Learning multi-view image-based rendering.
\newblock In {\em CVPR}, 2021.

\bibitem{weng2015convolutional}
W Weng and X Zhu.
\newblock Convolutional networks for biomedical image segmentation.
\newblock {\em IEEE Access}, 2015.

\bibitem{wiles2020synsin}
Olivia Wiles, Georgia Gkioxari, Richard Szeliski, and Justin Johnson.
\newblock Synsin: End-to-end view synthesis from a single image.
\newblock In {\em Proceedings of the IEEE/CVF Conference on Computer Vision and
  Pattern Recognition}, pages 7467--7477, 2020.

\bibitem{wu20153d}
Zhirong Wu, Shuran Song, Aditya Khosla, Fisher Yu, Linguang Zhang, Xiaoou Tang,
  and Jianxiong Xiao.
\newblock 3d shapenets: A deep representation for volumetric shapes.
\newblock In {\em Proc.~CVPR}, 2015.

\bibitem{xiang2021neutex}
Fanbo Xiang, Zexiang Xu, Milos Hasan, Yannick Hold-Geoffroy, Kalyan Sunkavalli,
  and Hao Su.
\newblock Neutex: Neural texture mapping for volumetric neural rendering.
\newblock In {\em Proceedings of the IEEE/CVF Conference on Computer Vision and
  Pattern Recognition}, pages 7119--7128, 2021.

\bibitem{xu2020grid}
Qiangeng Xu, Xudong Sun, Cho-Ying Wu, Panqu Wang, and Ulrich Neumann.
\newblock Grid-gcn for fast and scalable point cloud learning.
\newblock In {\em Proceedings of the IEEE/CVF Conference on Computer Vision and
  Pattern Recognition}, pages 5661--5670, 2020.

\bibitem{yao2018mvsnet}
Yao Yao, Zixin Luo, Shiwei Li, Tian Fang, and Long Quan.
\newblock {MVS}net: Depth inference for unstructured multi-view stereo.
\newblock In {\em Proc.~ECCV}, pages 767--783, 2018.

\bibitem{yariv2020multiview}
Lior Yariv, Yoni Kasten, Dror Moran, Meirav Galun, Matan Atzmon, Basri Ronen,
  and Yaron Lipman.
\newblock Multiview neural surface reconstruction by disentangling geometry and
  appearance.
\newblock In {\em Proc.~NeurIPS}, 2020.

\bibitem{yu2021plenoxels}
Alex Yu, Sara Fridovich-Keil, Matthew Tancik, Qinhong Chen, Benjamin Recht, and
  Angjoo Kanazawa.
\newblock Plenoxels: Radiance fields without neural networks.
\newblock {\em arXiv preprint arXiv:2112.05131}, 2021.

\bibitem{yu2021plenoctrees}
Alex Yu, Ruilong Li, Matthew Tancik, Hao Li, Ren Ng, and Angjoo Kanazawa.
\newblock Plenoctrees for real-time rendering of neural radiance fields.
\newblock {\em arXiv preprint arXiv:2103.14024}, 2021.

\bibitem{yu2020pixelnerf}
Alex Yu, Vickie Ye, Matthew Tancik, and Angjoo Kanazawa.
\newblock pixelnerf: Neural radiance fields from one or few images.
\newblock In {\em CVPR}, 2021.

\bibitem{zhang2020nerf++}
Kai Zhang, Gernot Riegler, Noah Snavely, and Vladlen Koltun.
\newblock Nerf++: Analyzing and improving neural radiance fields.
\newblock {\em arXiv preprint arXiv:2010.07492}, 2020.

\bibitem{zhang2018perceptual}
Richard Zhang, Phillip Isola, Alexei~A Efros, Eli Shechtman, and Oliver Wang.
\newblock The unreasonable effectiveness of deep features as a perceptual
  metric.
\newblock In {\em CVPR}, 2018.

\bibitem{zhou2014color}
Qian-Yi Zhou and Vladlen Koltun.
\newblock Color map optimization for {3D} reconstruction with consumer depth
  cameras.
\newblock {\em ACM Transactions on Graphics}, 33(4):155, 2014.

\bibitem{zhou2018stereo}
Tinghui Zhou, Richard Tucker, John Flynn, Graham Fyffe, and Noah Snavely.
\newblock Stereo magnification: learning view synthesis using multiplane
  images.
\newblock {\em ACM Transactions on Graphics}, 37(4):1--12, 2018.

\end{thebibliography}
}

\begin{appendices}
    \twocolumn[{%
        \renewcommand\twocolumn[1][]{#1}%
        \begin{center}
            \centering
            \LARGE \textbf{\appendixname}
            \vspace{30pt}
        \end{center}%
    }]
    
    \section{Ablation Studies on Point Features Initialization}
    \begin{table}[h]
        \begin{adjustwidth}{0pt}{0pt}  
        \centering
        \captionsetup{aboveskip=5pt}
        \setlength\tabcolsep{6pt}
        {\small
            \begin{tabular}{l|cc|cc}
            \hline
            & \multicolumn{1}{l}{Extract$_{20k}$} & \multicolumn{1}{l|}{Rand$_{20k}$} & \multicolumn{1}{l}{Extract$_{200k}$} & \multicolumn{1}{l}{Rand$_{200k}$} \\ \hline
            PSNR$\uparrow$     & \textbf{30.09}                & 25.44                          & \textbf{33.00}                           & 32.01                          \\
            SSIM$\uparrow$  & \textbf{0.963}                & 0.932                          & \textbf{0.978}                           & 0.972                         \\ \hline
            \end{tabular}
        }
        \caption {Comparisons between using the extracted image features to initialize the point features (our full model) or using the random initialized features.}
        \label{tb:ab_init} 
        \end{adjustwidth}
    \end{table}
    We conduct experiments to demonstrate the importance of our feature initialization. We compare our full model and our model initialized without using the extracted image features on the NeRF Synthetic dataset \cite{mildenhall2020nerf}. Without using the features from images, we randomly initialize the point features by using the popular Kaiming Initialization \cite{he2015delving}. As shown in Table \ref{tb:ab_init}, the neural points with image features not only achieve better performance after convergence at $200K$ iterations but also converge much faster in the beginning. The randomly initialized neural points even cannot perform as well as our full model, still outperforms state-of-the-art methods such as NeRF and NSVF\cite{liu2020neural}.

    \section{Per-scene Breakdown Results of the DTU Dataset}
    \begin{table}[]
      \setlength\tabcolsep{6pt}
      \captionsetup{aboveskip=5pt}
      \small{
          \begin{tabular}{lccccc}
            \hline
            \multicolumn{1}{l|}{Scan}       & \#1                  & \#8   & \#21                 & \#103                & \#114                \\ \hline
                        \multicolumn{6}{c}{SSIM$\uparrow$} \\ \hline
            \multicolumn{1}{l|}{Ours$_{1K}$}     & 0.935                & 0.906 & 0.913                & 0.944                & 0.948                \\
            \multicolumn{1}{l|}{Ours$_{10K}$}    & 0.962                & 0.949 & 0.954                & 0.961                & 0.960                \\
            \multicolumn{1}{l|}{MVSNeRF$_{10K}$\cite{chen2021mvsnerf}} & 0.934                & 0.900 & 0.922                & 0.964                & 0.945                \\
            \multicolumn{1}{l|}{IBRNET$_{10K}$\cite{ibrnet}}  & 0.955                & 0.945 & 0.947                & 0.968                & 0.964                \\
            \multicolumn{1}{l|}{NeRF$_{200K}$\cite{mildenhall2020nerf}} & 0.902                & 0.876 & 0.874                & 0.944                & 0.913                \\ \hline
                    \multicolumn{6}{c} {LPIPS$_{Vgg}\downarrow$} \\ \hline 
            \multicolumn{1}{l|}{Ours$_{1K}$}     & 0.151                & 0.207 & 0.201                & 0.208                & 0.148                \\
            \multicolumn{1}{l|}{Ours$_{10K}$}    & 0.095                & 0.130 & 0.134                & 0.145                & 0.096                \\
            \multicolumn{1}{l|}{MVSNeRF$_{10K}$} & 0.171                & 0.261 & 0.142                & 0.170                & 0.153                \\
            \multicolumn{1}{l|}{IBRNET$_{10K}$}  & 0.129                & 0.170 & 0.104                & 0.156                & 0.099                \\
            \multicolumn{1}{l|}{NeRF$_{200K}$}   & 0.265                & 0.321 & 0.246                & 0.256                & 0.225                \\ \hline
                        \multicolumn{6}{c} {PSNR$\uparrow$} \\ \hline 
            \multicolumn{1}{l|}{Ours$_{1K}$}     & 28.79                & 28.39 & 24.78                & 30.36                & 29.82                \\
            \multicolumn{1}{l|}{Ours$_{10K}$}    & 30.85                & 30.72 & 26.22                & 32.08                & 30.75                \\
            \multicolumn{1}{l|}{MVSNeRF$_{10K}$} & 28.05                & 28.88 & 24.87                & 32.23                & 28.47                \\
            \multicolumn{1}{l|}{IBRNET$_{10K}$}  & 31.00                & 32.46 & 27.88                & 34.40                & 31.00                \\
            \multicolumn{1}{l|}{NeRF$_{200K}$}   & 26.62                & 28.33 & 23.24                & 30.40                & 26.47                \\ \hline
            \end{tabular}           
        }
        \caption{Quantity comparison on five sample scenes in the DTU testing set with the view synthesis setting introduced in \cite{chen2021mvsnerf}. The subscripts indicate the number of iterations during optimization.}
        \label{tb:dt_dtu}
    \end{table}
    We show the per scene detailed quantitative results of the comparisons on the DTU\cite{dtu} dataset in Table \ref{tb:dt_dtu} and additional qualitative comparisons in our video. Since our method also faithfully reconstructs the scene geometry, our method has the best SSIM scores in most of the cases. Our model also has the best LPIPS for most of the scenes and therefore, is more visually authentic, as shown in the Figure 6 of the main paper and the video. IBRNet combines the colors from the source views to compute the radiance colors during shading. This image-based approach results in better PSNR. However, as shown in our video, our method is more temporal consistent because the local radiance and geometries are consistently stored at each neural point location.
    
    \section{Per-scene Breakdown Results of the NeRF Synthetic Dataset}
    \begin{table*}[]
      \captionsetup{aboveskip=5pt}
      \centering
      \begin{tabular}{lcccccccc}
            \hline
            \multicolumn{9}{c}{NeRF Synthetic}               \\
                       & Chair          & Drums          & Lego           & Mic            & Materials      & Ship           & Hotdog         & Ficus          \\ \hline
            \multicolumn{9}{c}{PSNR$\uparrow$}                                                                                                                           \\ \hline
            NPBG\cite{aliev2020neural}       & 26.47          & 21.53          & 24.84          & 26.62          & 21.58          & 21.83          & 29.01          & 24.60          \\
            NeRF\cite{mildenhall2020nerf}       & 33.00          & 25.01          & 32.54          & 32.91          & 29.62          & 28.65          & 36.18          & 30.13          \\
            NSVF\cite{liu2020neural}       & 33.19          & 25.18          & 32.54          & 34.27          & \textbf{32.68} & 27.93          & 37.14 & 31.23          \\
            Point-NeRF$^{col}_{200K}$  & 35.09   &    25.01  &   32.65  &   35.54   &    26.97  &   30.18  &  35.49   &   33.24          \\
            Point-NeRF$_{20K}$  & 32.50 &        25.03 &        32.40 &        32.31 &        28.11 &        28.13 &        34.53 &        32.67           \\
            Point-NeRF$_{200K}$ & \textbf{35.40} &        \textbf{26.06} &        \textbf{35.04} &        \textbf{35.95} &        29.61 &        \textbf{30.97} &        \textbf{37.30} &        \textbf{36.13} \\ \hline
            \multicolumn{9}{c}{SSIM$\uparrow$}                                                                                                                           \\ \hline
            NPBG       & 0.939          & 0.904          & 0.923          & 0.959          & 0.887          & 0.866          & 0.964          & 0.940          \\
            NeRF       & 0.967          & 0.925          & 0.961          & 0.980          & 0.949          & 0.856          & 0.974          & 0.964          \\
            NSVF       & 0.968          & 0.931          & 0.960          & 0.987          & \textbf{0.973}          & 0.854          & 0.980          & 0.973          \\
            Point-NeRF$^{col}_{200K}$  & 0.990 &	0.944 &	0.983 &	0.993 &	0.955 &	0.941 &	0.986 &	0.989           \\
            Point-NeRF$_{20K}$  & 0.981 &        0.944 &        0.980 &        0.986 &        0.959 &        0.916 &        0.983 &        0.986          \\
            Point-NeRF$_{200K}$ & \textbf{0.991} &        \textbf{0.954} &        \textbf{0.988} &        \textbf{0.994} &        0.971 &        \textbf{0.942} &        \textbf{0.991} &        \textbf{0.993} \\ \hline

            \multicolumn{9}{c}{SSIM~(Calibrated)~$\uparrow$}   \\                       
            Point-NeRF$_{200K}$ & 0.984 &     0.935 &        0.978 &        0.990 &        0.948 &       0.892 &        0.982 &        0.987 \\ \hline
            \multicolumn{9}{c}{LPIPS$_{Vgg}\downarrow$}                                                                                                                       \\ \hline
            NPBG       & 0.085          & 0.112          & 0.119          & 0.060          & 0.134          & 0.210          & 0.075          & 0.078          \\
            NeRF       & 0.046          & 0.091          & 0.050          & 0.028          & 0.063          & 0.206          & 0.121          & 0.044          \\
            Point-NeRF$^{col}_{200K}$  & 0.026 &	0.099 &	0.031 &	0.019 &	0.100 &	0.134 &	0.061 &	0.028          \\
            Point-NeRF$_{20K}$  & 0.051 &        0.103 &        0.054 &        0.039 &        0.102 &        0.181 &        0.074 &        0.043          \\
            Point-NeRF$_{200K}$ & \textbf{0.023} &   \textbf{0.078} &  \textbf{0.024} &   \textbf{0.014} &   \textbf{0.072} &    \textbf{0.124} &        \textbf{0.037} &        \textbf{0.022} \\ \hline
            \multicolumn{9}{c}{LPIPS$_{Alex}\downarrow$}                     \\ \hline
            NSVF       & 0.043          & 0.069          & 0.029          & 0.010          & \textbf{0.021} & 0.162          & 0.025          & 0.017          \\
            Point-NeRF$^{col}_{200K}$  & 0.013 &       0.073 &        0.016 &        0.011 &        0.076 &        0.087 &        0.032 &        0.012          \\
            Point-NeRF$_{20K}$  & 0.027 &        0.057 &        0.022 &        0.024 &        0.076 &        0.127 &        0.044 &        0.022          \\
            Point-NeRF$_{200K}$ & \textbf{0.010} &   \textbf{0.055} &   \textbf{0.011} &   \textbf{0.007} &        0.041 &    \textbf{0.070} &    \textbf{0.016} &   \textbf{0.009} \\ \hline
        \end{tabular}      
        \caption{Detailed breakdown of quantitative metrics of individual scenes for the NeRF Synthetic \cite{mildenhall2020nerf} for our method and baselines. All scores are averaged over the testing images. The subscripts are the number of iterations of the models and Point-NeRF$^{col}_{200K}$ indicates our method initiates from COLMAP points and optimized for 200 thousand iterations. We have noticed the more recent studies used different SSIM setting during computation, therefore we also calibrated our SSIM for fair comparison.}
        \label{tb:dt_nerfsynth}
    \end{table*}
    We show the per scene detailed quantitative results of the comparisons on the NeRF Synthetic\cite{mildenhall2020nerf} dataset in Table \ref{tb:dt_nerfsynth} and additional qualitative comparisons in our video. Point-NeRF achieves the best PSNRs, SSIMs and LPIPSs on most of the scenes and outperforms state-of-the-art methods \cite{aliev2020neural,mildenhall2020nerf,liu2020neural,ibrnet} with a big margin. On the other hand, our method initiated with COLMAP points is on par with NeRF. Even starting from the unideal initial points, we still manage to improve the geometry reconstruction and generate a high-quality radiance field with point pruning and growing. The fact that our model at $20K$ iterations matches the results of NeRF at $500K$ iterations clearly demonstrates our ability of fast convergence. 
    
    \section{Evaluation on Large-scale 3D Scenes (ScanNet).} 
    \label{sec:scannet}
    \begin{table*}[]
      \setlength\tabcolsep{3pt}
      \centering
      \captionsetup{aboveskip=5pt}
        \begin{tabular}{lcccccccc}
         & \multicolumn{4}{c}{\textbf{Average over two scenes}}       & \textbf{Scene 101} & \textbf{Scene 241} & \textbf{Scene 101} & \textbf{Scene 241}\\ \hline
        \multicolumn{1}{l|}{}            & SRN~\cite{sitzmann2019scene}   & NeRF~\cite{martin2021nerf}  & NSVF~\cite{liu2020neural}  & \multicolumn{1}{c|}{Ours$_{300K}^{depth}$} & \multicolumn{2}{c|}{Ours$_{300K}^{depth}$}  & \multicolumn{2}{c} {Ours$_{100K}^{mesh}$}   \\ \hline
        \multicolumn{1}{l|}{PSNR~$\uparrow$}        & 18.25  & 22.99 & 25.48 & \multicolumn{1}{c|}{\textbf{30.32}}      & 30.13          & 30.51   & 21.98 & 29.86       \\
        \multicolumn{1}{l|}{SSIM ~$\uparrow$}       & 0.592  & 0.620 & 0.688 & \multicolumn{1}{c|}{\textbf{0.909 / 0.814}}      & 0.912 / 0.821        & 0.906 / 0.807   & 0.882 / 0.797 & 0.901 / 0.784      \\
        \multicolumn{1}{l|}{RMSE~$\downarrow$}        & 14.764 & 0.681 & 0.079 & \multicolumn{1}{c|}{\textbf{0.031}}      & 0.032          & 0.030     & 0.091 & 0.033      \\
        \multicolumn{1}{l|}{LPIPS$_{Alex}\downarrow$} & 0.586  & 0.369 & 0.301 & \multicolumn{1}{c|}{\textbf{0.220}}     & 0.203          & 0.238    & 0.283 & 0.263       \\
        \multicolumn{1}{l|}{LPIPS$_{Vgg}\downarrow$}  &   -   &   -   &   -   & \multicolumn{1}{c|}{\textbf{0.292}}      & 0.286          & 0.299     & 0.345 & 0.327      \\ \hline
        \end{tabular}
        \caption{Quantity comparison on two scenes in the ScanNet dataset \cite{dai2017scannet} selected in NSVF \cite{liu2020neural}. RMSE is the Root Mean Square Error. Our method Point-NeRF outperforms all state-of-the-art methods in all metrics by substantial margins. We also report the (original/calibrated) SSIM. Our original SSIM uses the Skimage library with dynamic max signal value while several current papers use SSIM with as 1, therefore we also calibrate to their settings. Besides, per other authors' requests, we also add our method starting with the mesh instead of depth images, and report its results after 100K steps. Since the mesh of Scene 101 is extremely incomplete, we can observe tremendous quality loss.}
        \label{tb:scannet}
    \end{table*}

    \begin{figure*}[]
        \begin{adjustwidth}{0pt}{0pt}
            \setlength{\abovedisplayskip}{0pt}%
            \setlength{\abovedisplayshortskip}{\abovedisplayskip}%
            \setlength{\belowdisplayskip}{0pt}%
            \begin{center}
                \begin{subfigure}{0.195\linewidth}
                    \includegraphics[width=1\linewidth,trim={10 10 10 10},clip]{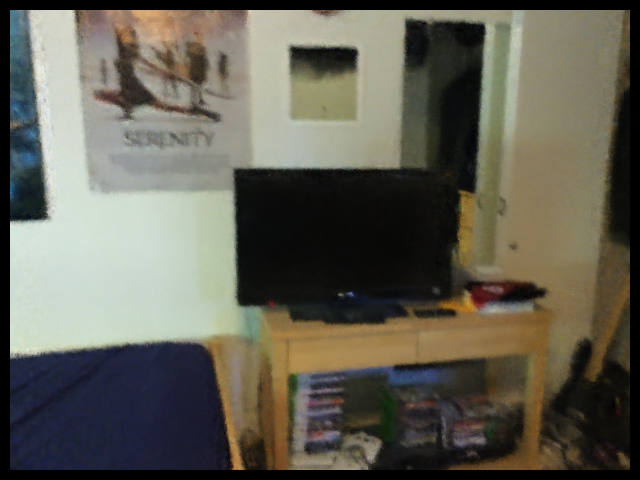}            
                    \captionsetup{aboveskip = 1pt}
                    \captionsetup{belowskip = 1pt}
                \end{subfigure}
                \begin{subfigure}{0.195\linewidth}
                    \includegraphics[width=1\linewidth,trim={10 10 10 10},clip]{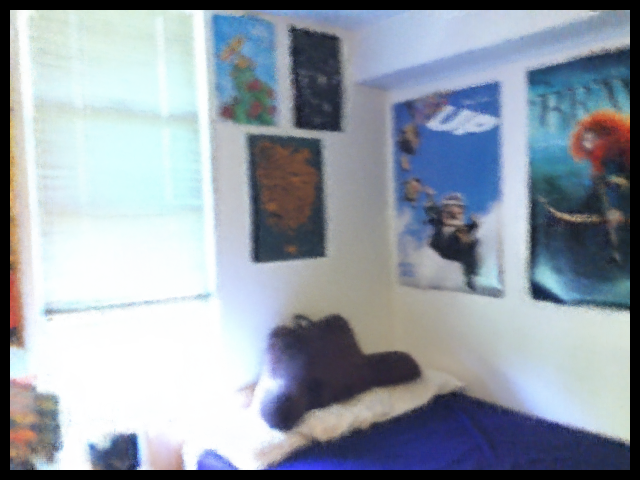}            
                    \captionsetup{aboveskip = 1pt}
                    \captionsetup{belowskip = 1pt}
                \end{subfigure}
                \begin{subfigure}{0.195\linewidth}
                    \includegraphics[width=1\linewidth,trim={10 10 10 10},clip]{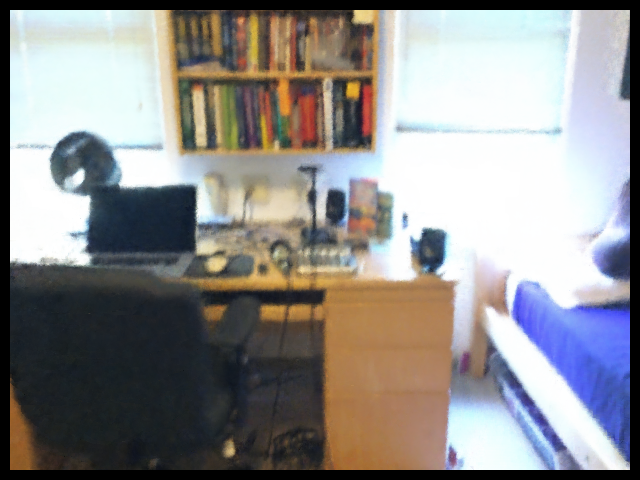}            
                    \captionsetup{aboveskip = 1pt}
                    \captionsetup{belowskip = 1pt}
                \end{subfigure}
                \begin{subfigure}{0.195\linewidth}
                    \includegraphics[width=1\linewidth,trim={10 10 10 10},clip]{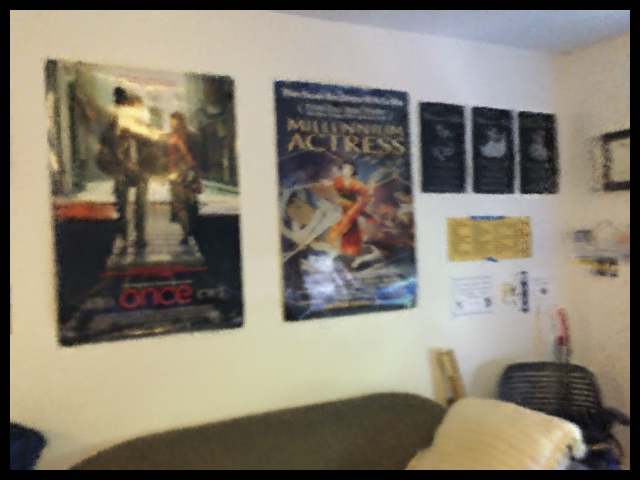}            
                    \captionsetup{aboveskip = 1pt}
                    \captionsetup{belowskip = 1pt}
                \end{subfigure}
                \begin{subfigure}{0.195\linewidth}
                    \includegraphics[width=1\linewidth,trim={10 10 10 10},clip]{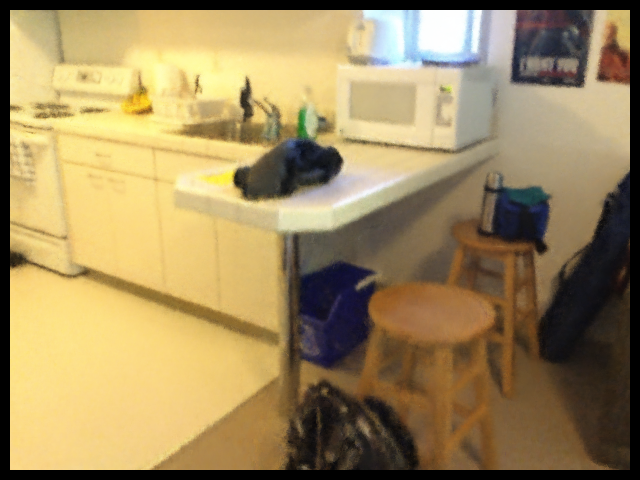}            
                    \captionsetup{aboveskip = 1pt}
                    \captionsetup{belowskip = 1pt}
                \end{subfigure}
                
                \begin{subfigure}{0.195\linewidth}
                    \includegraphics[width=1\linewidth,trim={10 10 10 10},clip]{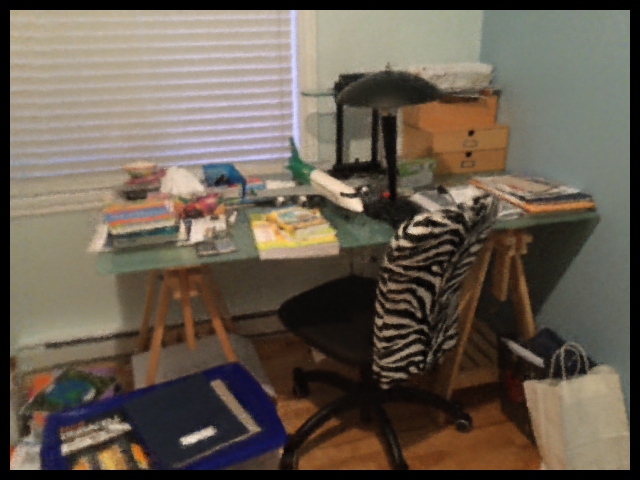}            
                    \captionsetup{aboveskip = 1pt}
                    \captionsetup{belowskip = 1pt}
                \end{subfigure}
                \begin{subfigure}{0.195\linewidth}
                    \includegraphics[width=1\linewidth,trim={10 10 10 10},clip]{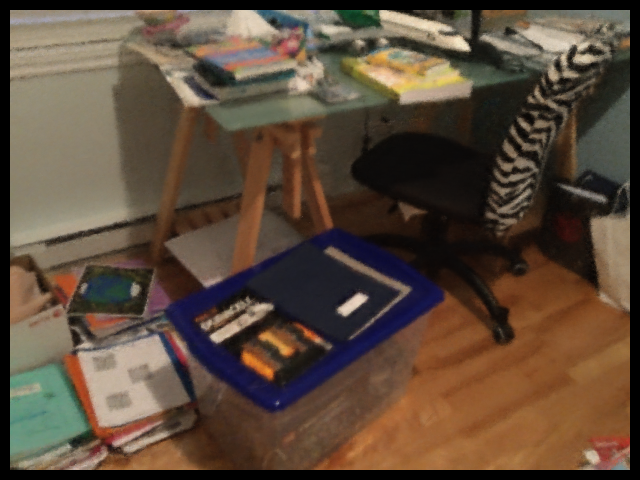}            
                    \captionsetup{aboveskip = 1pt}
                    \captionsetup{belowskip = 1pt}
                \end{subfigure}
                \begin{subfigure}{0.195\linewidth}
                    \includegraphics[width=1\linewidth,trim={10 10 10 10},clip]{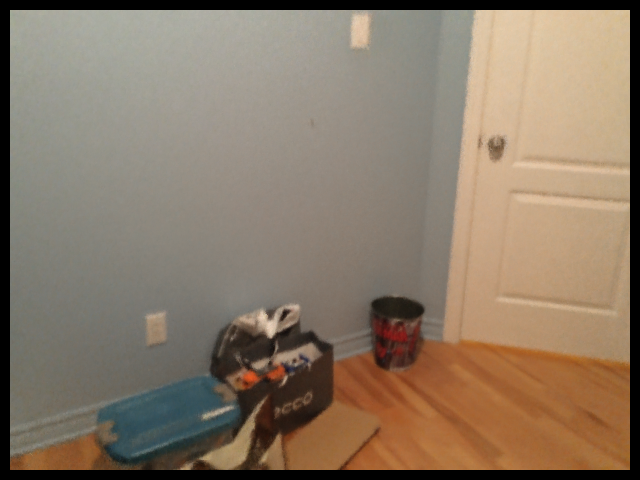}            
                    \captionsetup{aboveskip = 1pt}
                    \captionsetup{belowskip = 1pt}
                \end{subfigure}
                \begin{subfigure}{0.195\linewidth}
                    \includegraphics[width=1\linewidth,trim={10 10 10 10},clip]{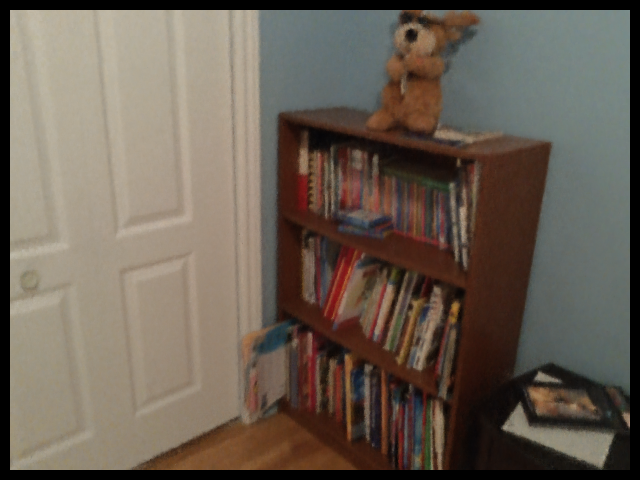}            
                    \captionsetup{aboveskip = 1pt}
                    \captionsetup{belowskip = 1pt}
                \end{subfigure}
                \begin{subfigure}{0.195\linewidth}
                    \includegraphics[width=1\linewidth,trim={10 10 10 10},clip]{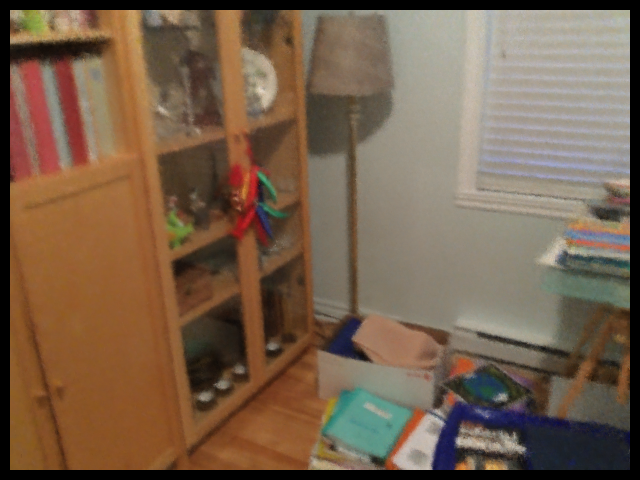}            
                    \captionsetup{aboveskip = 1pt}
                    \captionsetup{belowskip = 1pt}
                \end{subfigure}
            \end{center}
        \end{adjustwidth}
        \captionsetup{aboveskip = 2pt}
        \caption{The qualitative results of our Point-NeRF on the ScanNet dataset \cite{Knapitsch2017}. The first row shows five generated test frames of scene 101 and the second row shows five generated test frames of scene 241.}
        \label{fig:scannet}
    \end{figure*}
    While our model is purely trained on a dataset of objects (the DTU dataset), our network generalizes well to large-scale 3D scene datasets. Following \cite{liu2020neural}, we use two 3D scenes, scene $0101\_04$ and scene $0241\_01$, from ScanNet \cite{dai2017scannet}. We extract both RGB and depth images from the original videos and from which we sample one out of five frames as training set and use the rest for testing. The RGB images are scaled to 640 × 480. We finetune each scene for 300K steps with point pruning and growing.
    
    We compare with 3 other state-of-the-art methods with quantitative results in Tab.~\ref{tb:nerfsynth}.
    In particular, we compare with a scene representation model (SRN) \cite{sitzmann2019scene}, NeRF~\cite{mildenhall2020nerf} and a sparse voxel-based neural radiance field, NSVF~\cite{liu2020neural}.
    The qualitative comparison is shown in Tab.~\ref{tb:scannet} and visual results are shown in Figure~\ref{fig:scannet}. Our Point-NeRF outperforms all these previous studies in all metrics by substantial margins. Please find more visual results in our video.
    
    \section{The Tanks and Temple Dataset}
    \begin{table*}[hbt!]
      \centering
      \captionsetup{aboveskip=5pt}
        \begin{tabular}{ccccccc}
        \hline
        \multicolumn{7}{c}{Tanks \& Tamples}                                                                                                                \\
        \multicolumn{1}{l}{} & Ignatius             & Truck                & Barn      & Caterpillar                 & Family               & Mean                 \\ \hline
        \multicolumn{1}{l}{} & \multicolumn{1}{l}{} & \multicolumn{1}{l}{} & PSNR~$\uparrow$      & \multicolumn{1}{l}{} & \multicolumn{1}{l}{} & \multicolumn{1}{l}{} \\ \hline
        NV~\cite{lombardi2019neural}                   & 26.54                & 21.71                & 20.82     & 20.71                & 28.72                & 23.70                \\
        NeRF~\cite{mildenhall2020nerf}                 & 25.43                & 25.36                & 24.05     & 23.75                & 30.29                & 25.78                \\
        NSVF~\cite{liu2020neural}                 & 27.91                & 26.92                & 27.16     & 26.44                & 33.58                & 28.40                \\
        Point-NeRF (Ours)          & \textbf{28.43}                & \textbf{28.22}                & \textbf{29.15}     & \textbf{27.00}                & \textbf{35.27}                & \textbf{29.61}                \\ \hline
        \multicolumn{1}{l}{} & \multicolumn{1}{l}{} & \multicolumn{1}{l}{} & SSIM~$\uparrow$      & \multicolumn{1}{l}{} & \multicolumn{1}{l}{} &                      \\ \hline
        NV~\cite{lombardi2019neural}                   & 0.992                & 0.793                & 0.721     & 0.819                & 0.916                & 0.848                 \\
        NeRF~\cite{mildenhall2020nerf}                 & 0.920                & 0.860                & 0.750     & 0.860                & 0.932                & 0.864                 \\
        NSVF~\cite{liu2020neural}                 & 0.930                & 0.895                & 0.823     & 0.900                & 0.954                & 0.900                 \\
        Point-NeRF (Ours)           & \textbf{0.961}                & \textbf{0.950}                & \textbf{0.937}     & \textbf{0.934}                & \textbf{0.986}                & \textbf{0.954}                 \\ \hline
        \multicolumn{1}{l}{} & \multicolumn{1}{l}{} & \multicolumn{1}{l}{} & LPIPS$_{Alex}\downarrow$ & \multicolumn{1}{l}{} & \multicolumn{1}{l}{} &                      \\ \hline
        NV~\cite{lombardi2019neural}                   & 0.117                & 0.312                & 0.479     & 0.280                & 0.111                & 0.260                 \\
        NeRF~\cite{mildenhall2020nerf}                 & 0.111                & 0.192                & 0.395     & 0.196                & 0.098                & 0.198                 \\
        NSVF~\cite{liu2020neural}                 & 0.106                & 0.148                & 0.307     & 0.141                & 0.063                & 0.153                 \\
        Point-NeRF (Ours)           & \textbf{0.069}                & \textbf{0.077}                & \textbf{0.120}     & \textbf{0.111}                & \textbf{0.024}                & \textbf{0.080}                 \\ \hline
        \multicolumn{1}{l}{} & \multicolumn{1}{l}{} & \multicolumn{1}{l}{} & LPIPS$_{Vgg}\downarrow$  & \multicolumn{1}{l}{} & \multicolumn{1}{l}{} &                      \\ \hline
        Point-NeRF (Ours)           & \textbf{0.079}                & \textbf{0.117}                & \textbf{0.180}     & \textbf{0.156}                & \textbf{0.046}                & \textbf{0.115}                 \\ \hline
        \end{tabular}
        \caption{Quantity comparison on five scenes in the Tanks and Temples dataset \cite{Knapitsch2017} selected in NSVF \cite{liu2020neural}. Our method Point-NeRF outperforms all state-of-the-art models in all metrics by substantial margins.}
        \label{tb:tt}
    \end{table*}
    
    \begin{figure*}[]
        \begin{adjustwidth}{0pt}{0pt}
            \setlength{\abovedisplayskip}{0pt}%
            \setlength{\abovedisplayshortskip}{\abovedisplayskip}%
            \setlength{\belowdisplayskip}{0pt}%
            \begin{center}
                \begin{subfigure}{0.25\linewidth}
                    \includegraphics[width=1\linewidth,trim={200 0 200 0},clip]{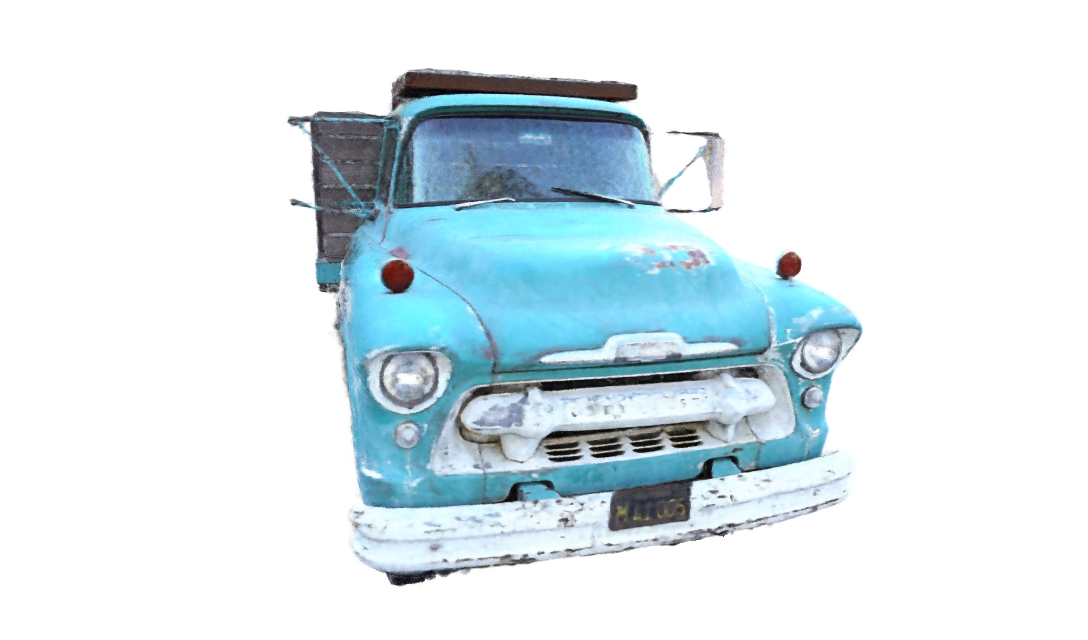}
                    \captionsetup{aboveskip = 1pt}
                    \captionsetup{belowskip = 1pt}
                \end{subfigure}
                \begin{subfigure}{0.23\linewidth}
                    \includegraphics[width=1\linewidth,trim={400 0 200 0},clip]{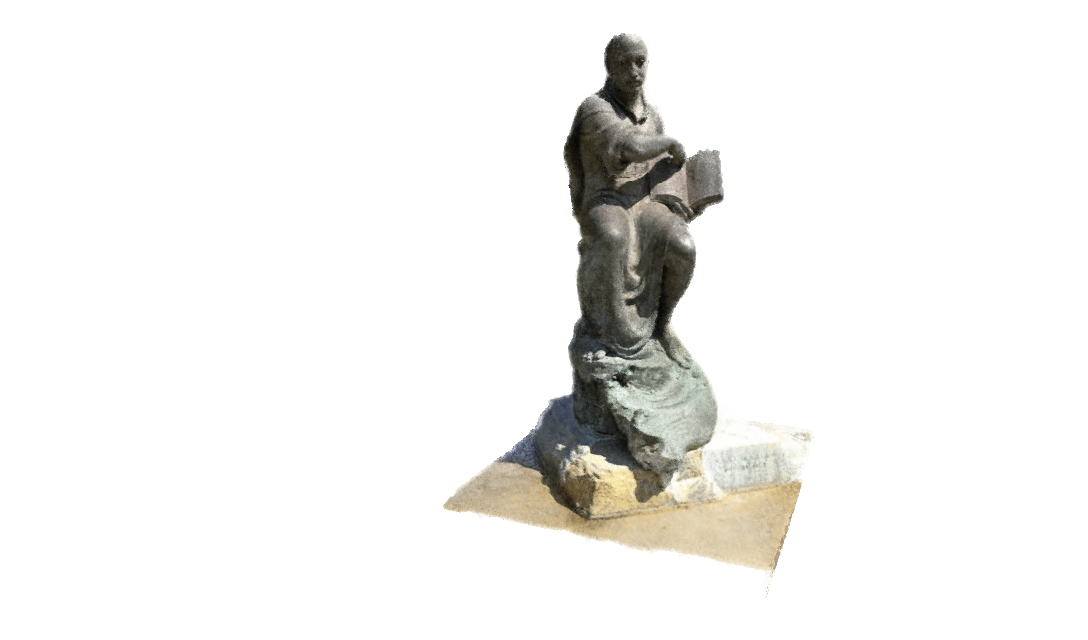}            
                    \captionsetup{aboveskip = 1pt}
                    \captionsetup{belowskip = 1pt}
                \end{subfigure}
                \begin{subfigure}{0.25\linewidth}
                    \includegraphics[width=1\linewidth,trim={0 0 0 0},clip]{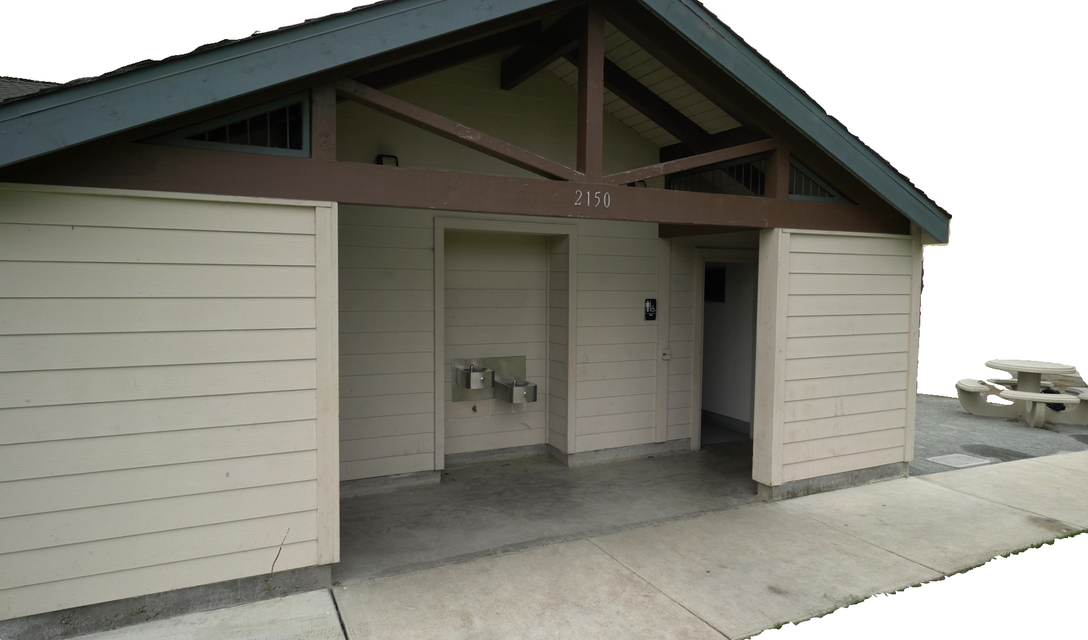} 
                    \captionsetup{aboveskip = 1pt}
                    \captionsetup{belowskip = 1pt}
                \end{subfigure}
                \begin{subfigure}{0.23\linewidth}
                    \includegraphics[width=1\linewidth,trim={150 0 350 0},clip]{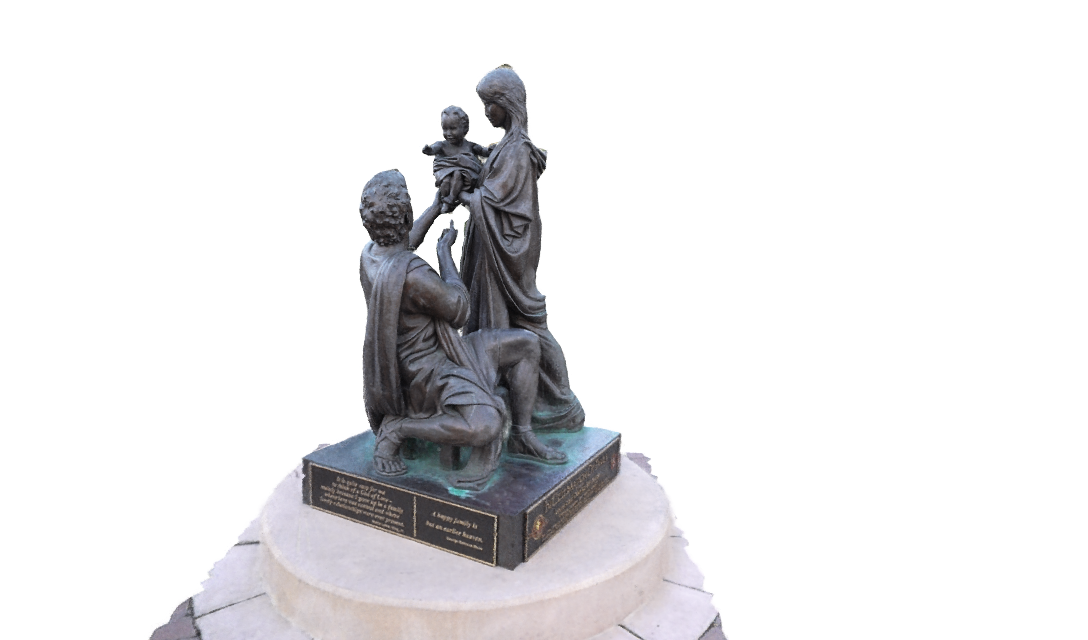} 
                    \captionsetup{aboveskip = 1pt}
                    \captionsetup{belowskip = 1pt}
                \end{subfigure}
            \end{center}
        \end{adjustwidth}
        \captionsetup{aboveskip = 2pt}
        \caption{The qualitative results of our Point-NeRF on the Tanks and Temples dataset.}
        \label{fig:tt}
    \end{figure*}
    We also experiment Point-NeRF on the Tanks and Temples dataset \cite{Knapitsch2017}. we reconstruct the radiance field of five scenes selected in NSVF~\cite{liu2020neural} and compare our model with three models NV \cite{lombardi2019neural}, NeRF \cite{mildenhall2020nerf} and NSVF \cite{liu2020neural}. We show the quantitative comparison in Tab.~\ref{tb:tt} and visualize quality results in Figure \ref{fig:tt}. Please find more visual results in our video.
    
    \section{Initializing Neural Points from COLMAP Points}
        Point-NeRF can use the points of any external reconstruction method. For instance, the output of COLMAP\cite{schoenberger2016mvs} is a point cloud $\{(\PointX_\iP)|\iP=1,...,\PointNum\}$. We set $\PointConf_\iP$ as $0.3$ in the beginning. The confidence score of valid points will be pushed to 1 during the optimization process. To acquire point features $\PointF_\iP$ for a point, We first rule out all the views where the point is occluded by other points, then we find the view of which the camera is the closest to the point. Then from that view, we can unproject the point onto the feature maps extracted by $G_f$ (see Figure 2(a) in the main paper) from the selected view and obtain the $\PointF_\iP$.
        
    \section{Networks Architectures}
         \begin{figure*}[]
            \begin{adjustwidth}{0pt}{0pt}
                \setlength{\abovedisplayskip}{0pt}%
                \setlength{\abovedisplayshortskip}{\abovedisplayskip}%
                \setlength{\belowdisplayskip}{0pt}%
                \begin{center}
                    \includegraphics[width=0.9\textwidth]{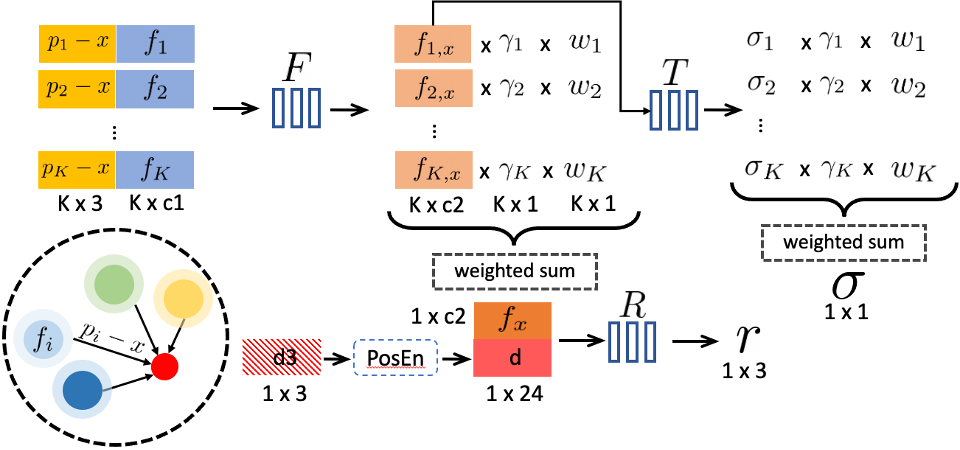}
                \end{center}
            \end{adjustwidth}
            \captionsetup{aboveskip = 2pt}
            \caption{The network pipeline of radiance fields computation at a shading location $x$ from $K$ neural points neighbors. ``PosEN'' indicates positional encoding \cite{mildenhall2020nerf}. ``d3'' indicates the 3 channels vector of view directions at $x$. The final outputs are the radiance color $r$ and density $\sigma$. Please also refer to the equations (3-7) in the main paper. }
            \label{fig:mlp}
        \end{figure*}
        \boldstartspace{Cost volume-based CNN $G_{p,\gamma}$.} Our cost volume-based CNN adopts the popular architecture of \cite{yao2018mvsnet}, which is simple and efficient. It includes three layers of depth features extraction CNN, while the latter two layers down-samples the spatial dimension by 4 and output a feature map with 32 channels. Then, these features from each view will be warped according to camera pose and the variance will be computed. The variance features will go through a narrow U-Net \cite{weng2015convolutional} and output a  1-channel feature to calculate the depth probability.  
        
        \boldstartspace{Image Feature Extraction 2D CNN $G_{f}$.} The image feature extraction network takes inputs of RGB image and has three down-sampling layers, each output feature with channels of $8, 16, 32$. We extract the point features by unprojecting a 3D point to each layer and taking the multi-scale features.
        
        \boldstartspace{Point-based Radiance Fields MLP.} We visualize the details of the point feature aggregation and radiance computation in Figure \ref{fig:mlp}. In all of our experiments, we set $c_1 = 56$, $c2 = 128$. The MLPs $F, R, T$ have 2, 3, 2 layers, respectively. The intermediate feature channels of $F$ and $T$ are 256, and 128 channels for $R$.

    \section{Neural Point Querying}
        To efficiently query neural point neighbors for ray marching, inspired by the CAGQ point query introduced in \cite{xu2020grid}, we implement a grid query method. 
        Then we build grid-point indices which register each neural point to evenly spaced 3D grids. Since these grids in the perspective coordinate are cubic, in the world coordinate, they have shapes of spherical voxels. 
        
        With the grid-point indices, we can discover grids that have neural points and also their grid neighbors. These grid neighbors are the regions of interest since there should exist neural points within the query radius. If a ray crosses these regions, we can place shading points inside. Finally, we query neural points by directly retrieving the stored neural points according to the grid-point indices.
        
        In all of our experiments, we query 8 nearest neural point neighbors for each shading location. Along each ray, we only search for neural point neighbors and compute radiance for shading locations in a grid that is occupied itself or nearby occupied grids. Therefore, our shading is much more efficient by skipping the empty space, unlike other radiance fields representations. This is one key advantage that enables fast convergence. Even NSVF \cite{liu2020neural}, high-performance local radiance representation, has to probe the empty space in the beginning and gradually prune the voxels along its training process.
        
        The benefit of this strategy is two-fold: First, we only place shading points in the area that exists neural points, so that we avoid radiance computation in the empty space. Second, the nearby points can be efficiently retrieved according to the indices, which substantially accelerate the point query speed.
        
    \section{Limitations}
    Because we do not focus on the rendering speed and we have not optimized our implementation (point querying and point feature aggregation) for fast rendering. Although, our model is naturally faster than NeRF (3X) due to that we skip the shading in empty space. We believe future works on combining mechanisms introduced in current papers such as \cite{yu2021plenoctrees,reiser2021kilonerf} with our point-based radiance representation would further benefit the neural rendering technology.

    \section{Additional Discussion and Issues Need Attention}
    \paragraph{Processing the points generated by MVSNet} We have received constructive feedbacks and hope to make it clear that when Point-NeRF uses MVSNet\cite{yao2018mvsnet} to reconstruct point cloud, the point fusion after depth estimation by MVSNet will use the alpha channel in the NeRF-Synthetic Dataset (as our published code indicates). It is due to the fact that MVSNet cannot handle background very well and will create too many outlier points in the background areas. Since images in the Tanks and Temples Dataset \cite{Knapitsch2017} don't have a alpha channel, we filter out the MVSNet points that appear in the regions of the pure background color. On the NeRF-Synthetic Dataset, the methods we compared with \cite{martin2021nerf,liu2020neural}, used the inputs: RGB images with the knowledge of the pure color background. Therefore, To improve the fairness, on the NeRF-Synthetic Dataset, we include results of Point-NeRF with MVSNet when using background color for filtering (not the alpha channel anymore). Its results is shown in Table \ref{tb:bg_nerfsynth} and one can cite which ever setting one thinks is fair. 
    
    Please note that, in our experiments, COLMAP doesn't use any filtering. Therefore, there is no impact on COLMAP results. When compare with NPGB \cite{aliev2020neural}, we use the same point cloud. Since it is more meaningful to rule out the impact of the point cloud quality, we advocate other point-based rendering works to use the same point cloud if willing to compare with our results. The point clouds are included in the checkpoints we published in the github repo.

    Our original intention of using MVSNet is due to its simplicity and the fact that it is one of the earlies deep learning-based MVS model. We, thus, encourage users to try a more advanced MVS model so that no filtering is needed.

    \paragraph{ScanNet and Unbounded Scenes}
    We also receive comments about our ScanNet experiments, and we would like to state very clearly that we use the depth images from the ScanNet Dataset to initialize the point cloud. It is because NSVF is our major baseline on this dataset and it uses this setting. In our original paragraph Appendix \ref{sec:scannet} we have provided this information, and we hope this could clear the potential false expectation from readers.

    Since Point-NeRF is a local radiance representation, without additional components, such as an additional background NeRF (used by Plenoxel \cite{yu2021plenoxels}), it cannot handle background in Unbounded scenes (also known as inside-out scenes). For ScanNet, there is not much of background since it is a indoor scene with noisy depth images, every parts in the room can be deemed as foreground. 

    \begin{table*}[]
      \captionsetup{aboveskip=5pt}
      \centering
      \begin{tabular}{lccccccccc}
            \multicolumn{10}{c}{Point-NeRF with MVSNet (background color filtering) on NeRF Synthetic} \\ \hline 
             & Chair & Drums & Lego  & Mic   & Materials & Ship  & Hotdog & Ficus & Mean  \\ \hline  
            PSNR~$\uparrow$  & 35.60 & 26.04 & 35.27 & 35.91 & 29.65     & 30.61 & 37.34  & 35.61 & 33.25 \\
            SSIM~$\uparrow$ & 0.991 & 0.954 & 0.989 & 0.994 & 0.971     & 0.938 & 0.991  & 0.992 & 0.978 \\
            LPIPS$_{Alex}\downarrow$ & 0.023 & 0.078 & 0.021 & 0.014 & 0.071 & 0.129 & 0.036  & 0.025 & 0.050 \\
            LPIPS$_{Vgg}\downarrow$ & 0.010 & 0.055 & 0.010 & 0.007 & 0.041 & 0.076 & 0.016  & 0.011 & 0.028 \\ \hline 
        \end{tabular}      
        \caption{We use MVSNet \cite{yao2018mvsnet} to reconstruct the points and filter them by using background color, then initialize neural points and optimized our Point-NeRF model for 200 thousand iterations.}
        \label{tb:bg_nerfsynth}
    \end{table*}

\end{appendices}

\end{document}